\documentclass{article}

\usepackage{microtype}
\usepackage{graphicx}
\usepackage{booktabs}
\usepackage{multirow}
\usepackage{makecell}
\usepackage{tabularx}
\usepackage{xcolor}
\usepackage{pifont}
\usepackage{forest}
\useforestlibrary{edges}
\usepackage{hyperref}

\newcommand{\prim}{\ensuremath{\blacksquare}}
\newcommand{\pmark}{\ensuremath{\triangle}}

\usepackage[preprint]{icml2026}

\usepackage{amsmath}
\usepackage{amssymb}
\usepackage{mathtools}
\usepackage{amsthm}
\usepackage[capitalize,noabbrev]{cleveref}

\theoremstyle{plain}
\newtheorem{observation}{Observation}
\crefname{observation}{Observation}{Observations}
\Crefname{observation}{Observation}{Observations}
\theoremstyle{plain}

\theoremstyle{definition}
\newtheorem{definition}{Definition}
\theoremstyle{remark}

\newcommand{\cmark}{\ding{51}}
\newcommand{\xmark}{\ding{55}}

\newcommand{\G}{\mathcal{G}}
\newcommand{\V}{\mathcal{V}}
\newcommand{\E}{\mathcal{E}}
\newcommand{\Mset}{\mathcal{M}}
\newcommand{\Nb}{\mathcal{N}}
\newcommand{\h}{\mathbf{h}}
\newcommand{\z}{\mathbf{z}}
\newcommand{\x}{\mathbf{x}}
\newcommand{\A}{\mathbf{A}}
\newcommand{\X}{\mathbf{X}}
\newcommand{\Hm}{\mathbf{H}}
\newcommand{\R}{\mathbb{R}}
\newcommand{\AGG}{\operatorname{AGG}}
\newcommand{\UPD}{\operatorname{UPD}}

\icmltitlerunning{Multimodal Graph Learning: Models, Benchmarks, and Frontiers}

\begin{document}

\twocolumn[
  \icmltitle{Multimodal Graph Learning:\\
             A Survey of Models, Benchmarks, and Frontiers}

  \icmlsetsymbol{equal}{*}

  \begin{icmlauthorlist}
    \icmlauthor{Rui Xue}{aff1}
    \icmlauthor{Tianfu Wu}{aff1}
  \end{icmlauthorlist}

  \icmlaffiliation{aff1}{Department of Electrical and Computer Engineering, North Carolina State University, Raleigh, NC}

  \icmlcorrespondingauthor{Rui Xue}{rxue@ncsu.edu}

  \icmlkeywords{Multimodal Graph Learning, Multimodal Attributed Graphs,
    Graph Neural Networks, Graph Foundation Models, Survey}

  \vskip 0.3in
]

\printAffiliationsAndNotice{}

\begin{abstract}
Multimodal attributed graphs (MAGs), in which nodes carry text, images, and other modalities while edges encode relations among entities, underlie e-commerce, social media, content platforms, and creative applications. Learning on MAGs has developed along two complemen- tary lines. \textit{Graph-centric methods} redesign how modalities are fused, propagated, and connected within message passing, progressing from early multimodal graph neural networks to modality- aware propagation and multimodal graph structure learning. \textit{Foundation-model-centric} methods instead pretrain transferable multimodal graph encoders or bring graph context into multimodal large language models for both prediction and generation. In parallel, several benchmarks have appeared with different datasets and evaluation protocols. This survey takes an algorithm-centric and evaluation-grounded view of the field and makes three contributions. \textit{Modeling}: we introduce a unified message-passing formulation and organize methods into five levels at which multi- modal learning and graph structure are integrated, namely the feature, propagation, topology, learn- ing, and foundation-model levels, covering both discriminative models and graph-conditioned gen- erative models; within this formulation we derive representative methods as special cases, compare their assumptions, expressiveness, and cost, and identify cross-modal sensitivity as a property that separates architectures which otherwise appear similar. \emph{Datasets and evaluation:} we review the benchmarks released in recent years, summarize their datasets with the metrics used for each task,
define model-independent descriptors that characterize datasets by label homophily, average degree, clustering, and edge provenance, and map each evaluation protocol to the baselines it uses. \textit{Frontiers}:  we connect multimodal graph learning to AI agents, a currently active research direction, covering both the multimodal graphs that agents perceive and remember and graph engineering, the practice of organizing agent workflows as explicit graphs; we then discuss open problems, concrete efforts for the community, and research directions, including when graph structure helps in the presence of strong multimodal foundation models. Together, these parts are intended to help researchers choose model designs that match the properties of their data, compare methods under consistent protocols, and identify where multimodal graph learning can have the greatest impact.

\end{abstract}

\section{Introduction}
\label{sec:intro}

Many real-world graphs are multimodal by construction. Products on e-commerce
platforms are linked by co-purchase relations and described by titles,
reviews, and images; posts on social platforms are organized into threads and
carry both text and pictures; short videos are connected by co-viewing behavior
and described by covers and captions; artworks are related through shared
artists, schools, and styles. Such data are naturally modeled as
\emph{multimodal attributed graphs} (MAGs), in which every node carries
attributes from several modalities and edges encode relations among entities
\citep{zhu2025mmgraph,yan2025magb,wan2026openmag}. Learning on MAGs, which we
refer to as \emph{multimodal graph learning} (MGL), has moved quickly from
recommendation-oriented multimodal graph neural networks (GNNs)
\citep{wei2019mmgcn,tao2020mgat} to general-purpose MAG backbones
\citep{zhang2026context,zuo2026rolemag,li2026lion}, multimodal graph structure
learning \citep{zhu2026tmte,xue2026recog}, multimodal graph foundation models
\citep{he2025unigraph2,wu2026smgfm,yang2026charm}, and multimodal large
language models (MLLMs) that consume graph structure
\citep{yoon2023mmgl,ning2025graph4mm,fan2026mlaga,sun2026mario}.

\textbf{Why multimodal graph learning?}
We see four reasons why learning on MAGs is more than the sum of multimodal
learning and graph learning.

\emph{Complementary semantics.} Text and images describe the same entity from
different perspectives. A product title states brand and function, while the
image conveys style and appearance; neither is a sufficient description on its
own. Benchmarks and recent methods show that jointly using textual and visual attributes can improve graph learning over either modality alone \citep{zhu2025mmgraph,yan2025magb,liu2026graphvlm,xue2026recog}.

\emph{Relational context.} The graph supplies information that does not exist
in any single sample. Co-purchase edges reveal complementary usage and
co-viewing edges reveal shared audiences; no amount of per-node encoding can
recover such information from the node's own attributes. This context also
benefits generation: conditioning image synthesis on a node's graph neighbors,
rather than on its text alone, yields images more consistent with the ground
truth \citep{jin2024instructg2i}.

\emph{Cross-modal relational reasoning.} The text of a neighbor can
disambiguate the image of the target node, and vice versa. Recent analyses show
that this cross-modal, cross-node interaction is not symmetric: when one
modality is degraded, the useful direction of completion is from the stronger
modality of the neighbors to the weaker modality of the target
\citep{zuo2026rolemag}. Exploiting such interactions requires architectures
that let information flow \emph{across modalities and across nodes} at the same
time.

\emph{Noisy and incomplete modalities and topology.} Real MAGs simultaneously
exhibit feature noise, modality imbalance, missing modalities (e.g., cold-start
items without images or isolated nodes without edges), and noisy or
task-agnostic edges \citep{hu2026ntsformer,zhu2026tmte,zhang2026context}. These
imperfections interact: a reliable edge can compensate for a missing modality,
and a reliable modality can reveal that an edge is spurious.

\textbf{Why neither side suffices alone.}
Multimodal foundation models such as CLIP \citep{radford2021clip} and
LLaVA-style MLLMs \citep{liu2023llava} typically align modalities in a pairwise
manner, one instance at a time, and overlook structural relations across data
points \citep{liu2026graphvlm}. Standard GNNs, conversely, treat node
attributes as a single feature vector and perform one shared message passing
over a single graph. This implicitly assumes that a neighbor is equally useful
for all modalities, an assumption that does not hold empirically: text and image
branches prefer different neighborhood organizations
\citep{zuo2026rolemag}. Consequently, the default recipe of
``encode each modality, concatenate, then run a GNN'' leaves both cross-modal
and structural signal unused~\citep{xue2026recog}, which motivates the designs surveyed in this
paper.

\subsection{Why a New Survey?}
\label{sec:intro:why}

Three prior works review learning with multimodal graphs.
\citet{ektefaie2023multimodal} propose a blueprint for graph-centric
multimodal learning with four components (identifying entities, uncovering
topology, propagating information, and mixing representations), and apply it to
image-intensive, language-intensive, and knowledge-intensive graphs.
\citet{peng2024learning} categorize multimodal graphs into feature-level,
node-level, and graph-level types, and categorize methods into multimodal graph
convolution (MGCN), multimodal graph attention (MGAT), and multimodal graph
contrastive learning (MGCL). \citet{wang2025mgllm} chart a roadmap toward
multimodal graph large language models (MG-LLMs) and articulate five desired
capabilities. These works established the vocabulary of the field. Since 2024,
however, the field has changed in three ways that call for a new synthesis.

\textbf{Gap 1: Recent models have outgrown existing taxonomies.}
Early multimodal graph models, which exemplify the categories used by prior surveys, fuse modalities statically: they either concatenate modality features before propagation or, as in MMGCN and MGAT, propagate each modality separately
and combine the results at the output. In neither case can the interaction between modalities adapt to the neighbors and the task during propagation. Recent methods instead fuse modalities dynamically,
through cross-modal coupled propagation, role-aware neighbor routing, and the co-evolution of topology and modality representations
\citep{zhang2026context,zuo2026rolemag,li2026lion,zhu2026tmte,xue2026recog}.
Neither backbone-based taxonomies (MGCN / MGAT / MGCL) nor fusion-stage taxonomies (early / intermediate / late) can distinguish these design choices.
At the same time, MLLMs fine-tuned on a specific MAG can reach state-of-the-art
results in many settings \citep{liu2026graphvlm}, yet well-designed graph-centric models remain competitive and also achieve state-of-the-art
performance \citep{xue2026recog,zhang2026context}. Neither line subsumes the other, so we cover graph-centric and foundation-model-centric methods in one view, organized by \emph{where} multimodal learning and graph structure are integrated
within a single formulation (\cref{sec:prelim,sec:taxonomy}).

\textbf{Gap 2: New benchmarks are not covered, and evaluation has fragmented.}
Earlier multimodal graph models were evaluated mainly on recommendation datasets
with ranking metrics, or on domain-specific graphs such as multimodal knowledge
graphs and biomedical networks with their own preprocessed features; prior
surveys accordingly list domain resources rather than general MAG benchmarks
\citep{peng2024learning}. Three developments motivated a new generation of
benchmarks: MAG backbones are now expected to serve general node- and edge-level
tasks beyond recommendation; the choice of feature encoder, from CLIP to MLLMs,
has become a variable that must be controlled; and modality-centric and
generative tasks require datasets with raw text and images. Five benchmarks
appeared within two years in response
\citep{zhu2025mmgraph,yan2025magb,liu2026graphvlm,wan2026openmag,li2026mm}.
They differ in datasets, encoders, splits, tasks, and baseline sets, and recent
method papers follow different ones, so a reported improvement is interpretable
only relative to its protocol. Existing surveys predate most of these benchmarks
and do not analyze their protocols. We review them, map each protocol to its baselines, and characterize the datasets (Section~\ref{sec:bench}).

\textbf{Gap 3: Existing surveys do not connect the field to current research trends.}
The research agendas of prior surveys were set before the field's recent shift.
The challenges identified by \citet{peng2024learning}, such as modality
imbalance, cross-modal alignment, temporal dynamics, and efficiency, remain
relevant but are framed for GNN-based models, and the MG-LLM roadmap
\citep{wang2025mgllm} centers on building a single multimodal graph LLM. Since
then, AI agents have become one of the most active research directions. Agents
increasingly rely on explicit graphs to organize planning, tool use, memory, and
coordination \citep{bei2025graphsagents}; as they become multimodal, the graphs they perceive and remember increasingly carry multiple modalities, while the workflow graphs that organize their computation, a practice termed graph engineering \citep{grapheng2026}, raise related questions about how information from different channels is combined. Existing surveys on multimodal graph learning predate these
developments and do not connect the field to agents. At the same time, strong
MLLMs make it necessary to ask when explicit structure is needed at all, a
question that recent benchmarks now allow us to study empirically. We therefore
connect multimodal graph learning to AI agents (\cref{sec:ge}) and discuss open problems, concrete efforts, and research directions (Section~\ref{sec:open}).


\cref{tab:survey-compare} summarizes how this survey relates to prior work.

\begin{table*}[t]
\centering
\caption{Comparison with prior surveys and perspective papers on multimodal
graph learning. \cmark: covered; \pmark: partially covered; \xmark: not
covered. The assessment reflects the main focus of each work.}
\label{tab:survey-compare}
\footnotesize
\setlength{\tabcolsep}{4pt}
\resizebox{\textwidth}{!}{%
\begin{tabular}{lcccccc}
\toprule
 & Organizing principle
 & \makecell{Coverage\\(method years)}
 & \makecell{Algorithm-level\\unified formulation}
 & \makecell{2024--2026\\MAG methods}
 & \makecell{Benchmark \&\\protocol synthesis}
 & \makecell{FM / MLLM\\methods} \\
\midrule
\citet{ektefaie2023multimodal} & Blueprint + domain       & $\leq$2022 & \pmark & \xmark & \xmark & \xmark \\
\citet{peng2024learning}       & Backbone operator + graph type & $\leq$2023 & \pmark & \xmark & \xmark & \xmark \\
\citet{wang2025mgllm}          & Desired MG-LLM capabilities & $\leq$2025 & \pmark & \pmark & \pmark & \cmark \\
\textbf{Ours}                  & Integration level (5 levels) & $\leq$2026 & \cmark & \cmark & \cmark & \cmark \\
\bottomrule
\end{tabular}%
}
\end{table*}

\subsection{Scope and Contributions}
\label{sec:intro:scope}

\textbf{Scope.}
We focus on MAGs whose nodes carry attributes from multiple modalities
(primarily text and images), with a topology that is either observed or learned,
and on both graph-centric tasks (node classification, link prediction, node
clustering) and modality-centric tasks (modality matching, cross-modal
retrieval, alignment, and graph-conditioned generation). Models that originated
in multimodal recommendation are included when they have become standard MAG
baselines. Multimodal knowledge graphs and biomedical multimodal graphs share
some techniques but have distinct problem formulations and dedicated surveys;
we mention them only briefly in \cref{sec:apps}.

\textbf{Contributions.}
Our contributions fall into three parts, which we expand into five specific
contributions (C1--C5). Modeling and datasets are the main parts; the
expansion places them in a broader context.

\noindent\textbf{Modeling.}
\begin{itemize}
  \item[\textbf{C1}] \textbf{A unified formulation.} We write multimodal graph
  learning as a single message-passing template in which each integration point
  is an explicit component, derive representative methods as special cases, and
  identify cross-modal sensitivity as a property that separates otherwise
  similar architectures (\cref{sec:prelim}).
  \item[\textbf{C2}] \textbf{An integration-level taxonomy.} We organize methods
  into five levels (feature, propagation, topology, learning, and
  foundation-model level), including graph-conditioned text and image
  generation, and for each level we analyze differences, connections,
  strengths, limitations, and complexity
  (\cref{sec:taxonomy,sec:feature,sec:prop,sec:topo,sec:learn,sec:fm}).
\end{itemize}

\noindent\textbf{Datasets and evaluation.}
\begin{itemize}
  \item[\textbf{C3}] \textbf{A map of benchmarks, datasets, and baselines.} We compare five recent benchmark platforms, summarize their datasets and task metrics, define model-independent dataset descriptors (label homophily, average degree, clustering, and edge provenance), and map each evaluation protocol to its baseline set (\cref{sec:bench}).
\end{itemize}

  

\noindent\textbf{Frontiers.}
\begin{itemize}
  \item[\textbf{C4}] \textbf{Connections to AI agents (current research trend).}
  AI agents increasingly rely on explicit graphs to organize planning, tool use, memory, and multi-agent coordination \citep{bei2025graphsagents}. We connect multimodal graph learning to two kinds of agent graphs: the environment and memory graphs that agents perceive and maintain, which can be modeled as MAGs when their nodes retain multiple modalities, and the workflow and communication graphs of graph engineering \citep{grapheng2026}, which are not necessarily MAGs but share structural principles with multimodal graph learning. We close with directions at this intersection (Section~\ref{sec:ge}).
  \item[\textbf{C5}] \textbf{Open problems, concrete efforts, and research directions.} Grounded in the evidence from current benchmarks, we identify
  open questions, including the expressiveness of cross-modal message passing and
  when graph structure helps in the presence of strong multimodal foundation
  models, and propose concrete efforts for the community and research directions suggested by recent methods (Section~\ref{sec:open}).
\end{itemize}

\textbf{Organization.}
\cref{sec:prelim} introduces notation, task families, and the unified
formulation. \cref{sec:taxonomy} presents the taxonomy.
\cref{sec:feature,sec:prop,sec:topo,sec:learn,sec:fm} review the five
integration levels. \cref{sec:cross} discusses robustness, efficiency, and
federated settings. Section~\ref{sec:bench} reviews benchmarks, datasets, dataset characteristics, and evaluation protocols. \cref{sec:apps} covers applications,
\cref{sec:ge} connects multimodal graph learning to AI agents, including graph
engineering,
\cref{sec:open} discusses open problems and future directions, and
\cref{sec:conclusion} concludes.

\section{Preliminaries}
\label{sec:prelim}

\subsection{Notation and Problem Setup}
\label{sec:prelim:notation}

We first define all symbols used throughout the paper.

\textbf{Graph.}
$\G=(\V,\E)$ denotes a graph with node set $\V$ and edge set
$\E\subseteq\V\times\V$; $N=|\V|$ is the number of nodes, and
$\A\in\{0,1\}^{N\times N}$ is the observed adjacency matrix, with
$\A_{ij}=1$ iff $(i,j)\in\E$. $\Nb(i)$ is the set of neighbors of node $i$
under $\A$, and $\G_i$ denotes the ego-subgraph centered at node $i$ (node $i$,
a sampled multi-hop neighborhood, and the edges among them).

\textbf{Modalities.}
$\Mset$ is the set of modalities (e.g., $\Mset=\{\mathrm{t},\mathrm{v}\}$ for
text and vision) and $M=|\Mset|$. $x_i^{m}$ is the raw attribute of node $i$ in
modality $m$ (a string, an image, etc.). Some nodes may lack some modalities;
$\Mset_i\subseteq\Mset$ denotes the modalities observed for node $i$.

\textbf{Encoders and representations.}
$E_m(\cdot;\theta_m)$ is the encoder of modality $m$ with parameters
$\theta_m$, which may be frozen, adapted with parameter-efficient fine-tuning, or
fully trained. $\X^{m}\in\R^{N\times d_m}$ stacks the encoded attributes
$\x_i^{m}=E_m(x_i^m;\theta_m)\in\R^{d_m}$. $\h_i^{m,(\ell)}\in\R^{d}$ is the
modality-$m$ hidden representation of node $i$ after $\ell$ layers, with
$\h_i^{m,(0)}=\mathbf{P}_m\x_i^m$, and $L$ is the number of layers. The
learnable projection $\mathbf{P}_m\in\R^{d\times d_m}$ maps the outputs of
different encoders, which have different dimensions $d_m$, into a common
$d$-dimensional space so that modalities can be fused and exchanged.

\textbf{Operators.}
$\Gamma_m$ is a \emph{structure operator} that returns the adjacency
$\tilde{\A}^{m}=\Gamma_m(\A,\{\Hm^{m'}\}_{m'\in\Mset})$ used by the modality-$m$
stream, where $\Hm^{m'}$ stacks the representations of modality $m'$;
$\Nb^{m}(i)$ is the neighbor set of $i$ under $\tilde{\A}^{m}$.
$\psi_{m\leftarrow m'}$ is a message function that maps a modality-$m'$
representation of a neighbor to a message for the modality-$m$ stream, and
$\alpha_{ij}^{m\leftarrow m'}\ge 0$ is the \emph{cross-modal coupling
coefficient} that weights this message for the edge $(i,j)$. $\AGG$ is a
permutation-invariant aggregator (sum, mean, max, or attention), and $\UPD_m$ is
the update function of modality $m$. $F$ is a modality fusion operator,
$\z_i$ is the fused representation of node $i$, and $R$ is a task head. $y_i$ is
the label of node $i$ when one exists.

\textbf{Foundation-model interface.}
$\mathcal{T}$ is a graph-to-token transformation that turns an ego-subgraph into
an input for a foundation model, for example by verbalizing it into text,
linearizing it into a sequence of multimodal tokens, or projecting graph
embeddings into soft tokens. $\Phi$ denotes a (multimodal) large language model
or another sequence model that consumes these tokens.

\begin{definition}[Multimodal attributed graph]
A multimodal attributed graph is a tuple
$\G=(\V,\E,\{x_i^m\}_{i\in\V,\,m\in\Mset_i})$ in which edges are unimodal
relations among entities and every node carries raw attributes from one or more
modalities.
\end{definition}

\subsection{Task Families}
\label{sec:prelim:tasks}

Following recent benchmarks
\citep{wan2026openmag}, we distinguish two task families.

\textit{Graph-centric tasks}
evaluate whether representations preserve structural, relational, and
class-discriminative information.
\emph{Node classification} predicts $y_i$ for unlabeled nodes;
\emph{link prediction} scores node pairs $(i,j)$ for the existence of an edge;
\emph{node clustering} partitions $\V$ without labels.

\textit{Modality-centric tasks}
evaluate whether representations capture fine-grained correspondences across
modalities. \emph{Modality matching} decides whether two modality views belong
to the same node; \emph{cross-modal retrieval} retrieves the counterpart of a
query in another modality; \emph{modality alignment} measures how well the
modality representations agree; and \emph{graph-conditioned generation}
produces a target modality of a node from its graph context, namely
graph-to-text (G2Text) and graph-to-image (G2Image) generation
\citep{yoon2023mmgl,jin2024instructg2i,fang2025graphgpto}.

\subsection{A Unified Formulation}
\label{sec:prelim:unified}

Using the notation of \cref{sec:prelim:notation}, a broad class of MAG models
can be written as $L$ layers of modality-aware message passing. Let
$\mathbf{c}_{ij}^{m,(\ell)}$ denote the message that neighbor $j$ sends to the
modality-$m$ stream of node $i$ at layer $\ell$. Then
\begin{align}
\mathbf{c}_{ij}^{m,(\ell)} &= \sum_{m'\in\Mset}
\alpha_{ij}^{m\leftarrow m'}\,
\psi_{m\leftarrow m'}\big(\h_j^{m',(\ell)}\big),\label{eq:unified-msg}\\
\h_i^{m,(\ell+1)} &= \UPD_m\Big(\h_i^{m,(\ell)},\,
\AGG_{j\in\Nb^{m}(i)}\mathbf{c}_{ij}^{m,(\ell)}\Big),\label{eq:unified-layer}
\end{align}
where the neighborhoods are induced by the structure operator,
\begin{equation}
\label{eq:unified-structure}
\tilde{\A}^{m} = \Gamma_m\big(\A, \{\Hm^{m'}\}_{m'\in\Mset}\big),
\end{equation}
followed by fusion and a task head,
\begin{equation}
\label{eq:unified-fusion}
\z_i = F\big(\{\h_i^{m,(L)}\}_{m\in\Mset}\big), \qquad
\hat{y}_i = R(\z_i).
\end{equation}
The parameters (including, optionally, the encoder parameters $\theta_m$) are
trained with an objective $\mathcal{L}$ that may combine task supervision with
alignment, contrastive, or reconstruction terms. Foundation-model-centric
methods replace the message-passing backbone with a sequence model,
\begin{equation}
\label{eq:unified-fm}
\hat{y}_i \;(\text{or a generated } \hat{x}_i^{m}) = \Phi\big(\mathcal{T}(\G_i)\big).
\end{equation}

\textbf{Integration levels.}
\cref{eq:unified-msg,eq:unified-layer,eq:unified-structure,eq:unified-fusion,eq:unified-fm}
expose five places where multimodal learning and graph structure can be
integrated. These five places define our taxonomy (\cref{sec:taxonomy}):
\begin{itemize}
  \item \textbf{Feature level} (\cref{sec:feature}): $F$ is applied at
  $\ell=0$, collapsing all modalities into one stream that a standard GNN then
  propagates.
  \item \textbf{Propagation level} (\cref{sec:prop}): the design of the
  message functions $\psi_{m\leftarrow m'}$ and the coupling coefficients
  $\alpha_{ij}^{m\leftarrow m'}$, i.e., whether and how modalities exchange
  information during propagation.
  \item \textbf{Topology level} (\cref{sec:topo}): the design of the structure
  operator $\Gamma_m$, i.e., how multimodal semantics reshape the graph.
  \item \textbf{Learning level} (\cref{sec:learn}): the objective
  $\mathcal{L}$ and training protocol, including graph-aware cross-modal
  alignment, self-supervision, and pretraining.
  \item \textbf{Foundation-model level} (\cref{sec:fm}): the interface
  $\mathcal{T}$ and the model $\Phi$ in \cref{eq:unified-fm}.
\end{itemize}

\subsection{What the Integration Levels Allow: Cross-Modal Sensitivity}
\label{sec:prelim:sensitivity}

The five levels above describe \emph{where} multimodal learning and graph
structure are integrated. To compare them, we ask what each choice allows a
model to do. We focus on the capability that motivated multimodal graph learning
in \cref{sec:intro}, cross-modal relational reasoning: using the other
modalities of neighbors to represent a node. For example, a product with an
uninformative title may be recognizable from the images of its co-purchased
products, but only if its text representation can be influenced by those images.
We formalize this capability below and use it throughout the survey to compare
methods within and across levels.

For fixed parameters, we call a model \emph{cross-modally insensitive} if the
modality-$m$ representation of any node does not depend on the modality-$m'$
attributes of any node, for all $m'\neq m$, before fusion.

\begin{observation}[Cross-modal sensitivity of modality-decoupled models]
\label{obs:jacobian}
Consider \cref{eq:unified-msg,eq:unified-layer,eq:unified-structure}. Suppose (i)
$\alpha_{ij}^{m\leftarrow m'}=0$ for all edges $(i,j)$ and all $m'\neq m$, and
(ii) the structure operator $\Gamma_m$ depends only on $\A$ and $\Hm^{m}$.
Then for every layer $\ell\le L$, all nodes $i,j$, and all $m'\neq m$,
\begin{equation}
\frac{\partial \h_i^{m,(\ell)}}{\partial \x_j^{m'}} = \mathbf{0}.
\end{equation}
\end{observation}

\begin{proof}[Proof sketch]
By induction on $\ell$. At $\ell=0$, $\h_i^{m,(0)}=\mathbf{P}_m\x_i^m$ does not
depend on $\x_j^{m'}$. If no $\h^{m,(\ell)}$ depends on $\x^{m'}$, then under (i)
the message entering $\h_i^{m,(\ell+1)}$ is a function of
$\{\h_j^{m,(\ell)}\}_{j\in\Nb^{m}(i)}$ only, and under (ii) the neighborhood
$\Nb^{m}(i)$ itself does not depend on $\x^{m'}$.
\end{proof}

\paragraph{How the integration levels relate to cross-modal sensitivity.}
\emph{Feature level:} early fusion is cross-modally sensitive, since every
message carries all modalities, but all modalities share one neighborhood and
one propagation (\cref{sec:feature:analysis}). \emph{Propagation level:}
modality-decoupled models such as MMGCN satisfy (i) and are therefore
insensitive before fusion, whereas coupled models relax (i) and let the other
modalities of neighbors enter every layer (\cref{sec:prop}); ReCoG \citep{xue2026recog} proves this result for decoupled and late-fusion architectures. \emph{Topology level:} building or refining the graph from
other modalities violates (ii) and opens a second route for cross-modal
interaction, even when propagation keeps modalities separate
(\cref{sec:topo:analysis}). \emph{Learning level:} alignment and contrastive
objectives change what representations encode but not the dependency structure
of the forward computation, so they act through whichever routes the
architecture provides (\cref{sec:learn}). \emph{Foundation-model level:}
sensitivity depends on whether the graph-to-token interface passes the other
modalities of neighbors to the model (\cref{sec:fm}). Cross-modal sensitivity is
therefore necessary but not sufficient for exploiting cross-modal relational
signal; \cref{sec:ge} extends it to agent systems through an interventional
analogue.
\section{Taxonomy Overview}
\label{sec:taxonomy}

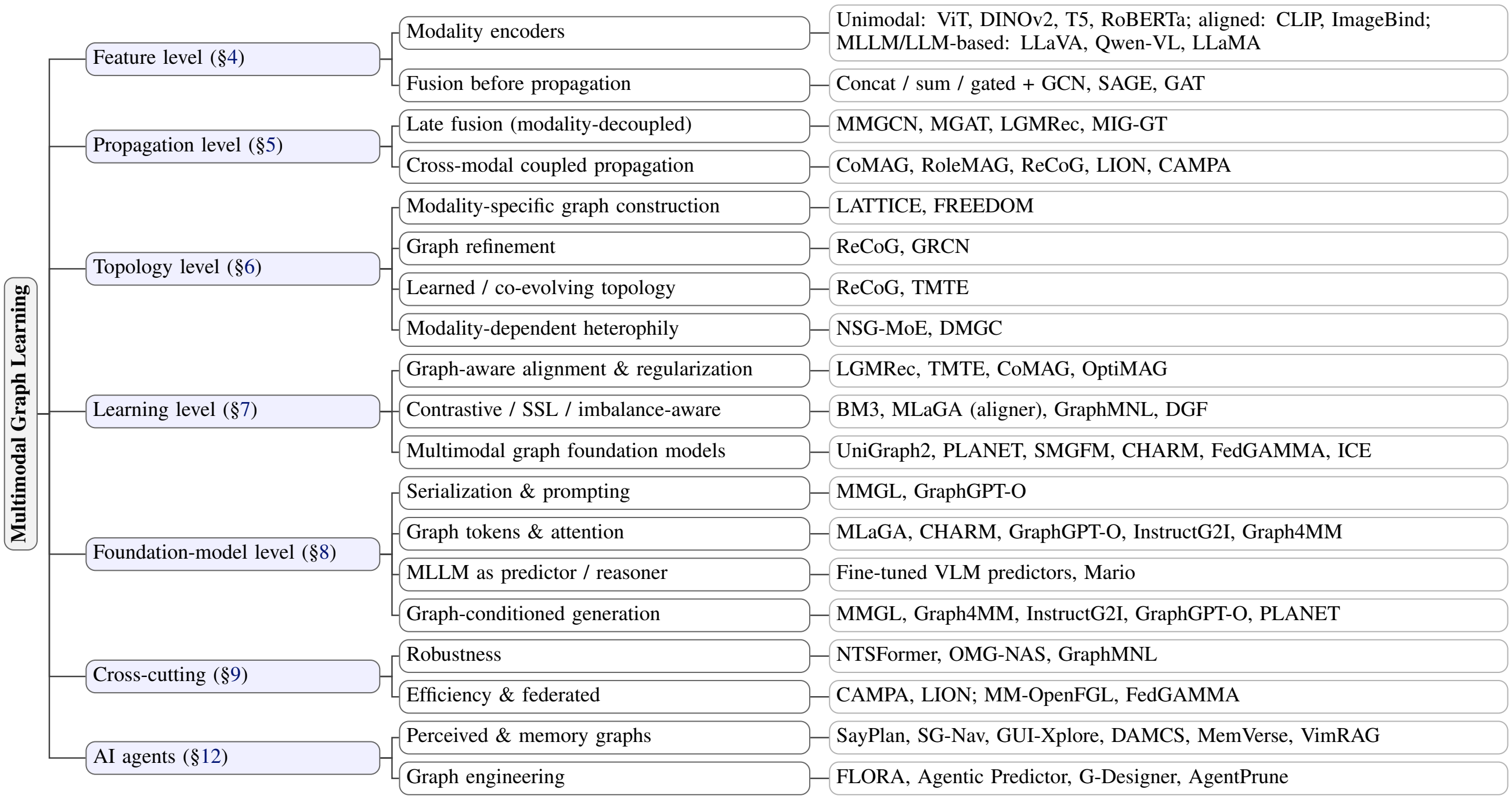
\begin{figure*}[t]
\centering
\resizebox{\textwidth}{!}{%
\begin{forest}
  forked edges,
  for tree={
    grow=east,
    reversed=true,
    anchor=base west,
    parent anchor=east,
    child anchor=west,
    base=left,
    font=\small,
    rectangle,
    draw=black!60,
    rounded corners,
    align=left,
    minimum width=4em,
    edge+={darkgray, line width=0.6pt},
    s sep=3pt,
    l sep=8pt,
    inner xsep=3pt,
    inner ysep=2pt,
    line width=0.5pt,
  },
  where level=0{font=\small\bfseries, fill=gray!10, align=center,
                rotate=90, anchor=center, parent anchor=south,
                child anchor=north}{},
  where level=1{text width=12em, font=\small, fill=blue!6}{},
  where level=2{text width=17em, font=\small}{},
  where level=3{text width=28.5em, font=\small, draw=black!30}{},
  [Multimodal Graph Learning
    [{Feature level (\S\ref{sec:feature})}
      [{Modality encoders}
        [{Unimodal: ViT, DINOv2, T5, RoBERTa; aligned: CLIP, ImageBind;\\ MLLM/LLM-based: LLaVA, Qwen-VL, LLaMA}]
      ]
      [{Fusion before propagation}
        [{Concat / sum / gated + GCN, SAGE, GAT}]
      ]
    ]
    [{Propagation level (\S\ref{sec:prop})}
      [{Late fusion (modality-decoupled)}
        [{MMGCN, MGAT, LGMRec, MIG-GT}]
      ]
      [{Cross-modal coupled propagation}
        [{CoMAG, RoleMAG, ReCoG, LION, CAMPA}]
      ]
    ]
    [{Topology level (\S\ref{sec:topo})}
      [{Modality-specific graph construction}
        [{LATTICE, FREEDOM}]
      ]
      [{Graph refinement}
        [{ReCoG, GRCN}]
      ]
      [{Learned / co-evolving topology}
        [{ReCoG, TMTE}]
      ]
      [{Modality-dependent heterophily}
        [{NSG-MoE, DMGC}]
      ]
    ]
    [{Learning level (\S\ref{sec:learn})}
      [{Graph-aware alignment \& regularization}
        [{LGMRec, TMTE, CoMAG, OptiMAG}]
      ]
      [{Contrastive / SSL / imbalance-aware}
        [{BM3, MLaGA (aligner), GraphMNL, DGF}]
      ]
      [{Multimodal graph foundation models}
        [{UniGraph2, PLANET, SMGFM, CHARM, FedGAMMA, ICE}]
      ]
    ]
    [{Foundation-model level (\S\ref{sec:fm})}
      [{Serialization \& prompting}
        [{MMGL, GraphGPT-O}]
      ]
      [{Graph tokens \& attention}
        [{MLaGA, CHARM, GraphGPT-O, InstructG2I, Graph4MM}]
      ]
      [{MLLM as predictor / reasoner}
        [{Fine-tuned VLM predictors, Mario}]
      ]
      [{Graph-conditioned generation}
        [{MMGL, Graph4MM, InstructG2I, GraphGPT-O, PLANET}]
      ]
    ]
    [{Cross-cutting (\S\ref{sec:cross})}
      [{Robustness}
        [{NTSFormer, OMG-NAS, GraphMNL}]
      ]
      [{Efficiency \& federated}
        [{CAMPA, LION; MM-OpenFGL, FedGAMMA}]
      ]
    ]
    [{AI agents (\S\ref{sec:ge})}
      [{Perceived \& memory graphs}
        [{SayPlan, SG-Nav, GUI-Xplore, DAMCS, MemVerse, VimRAG}]
      ]
      [{Graph engineering}
        [{FLORA, Agentic Predictor, G-Designer, AgentPrune}]
      ]
    ]
  ]
\end{forest}%
}
\caption{Taxonomy of multimodal graph learning by integration level. Methods may appear at multiple integration levels. See Table~\ref{tab:methods} for the full multi-label assignment.}
\label{fig:taxonomy}
\end{figure*}

\cref{fig:taxonomy} organizes the literature by the level at which
multimodal learning and graph structure are integrated, following the five
components identified in \cref{sec:prelim:unified}. Many recent methods
intervene at more than one level; for instance, ReCoG \citep{xue2026recog} exchanges messages between text and vision streams across
neighbors (propagation level) and learns the graph with a structure refiner
trained end to end with these streams (topology level). We therefore place each method under
its \emph{primary} contribution in \cref{fig:taxonomy} and record all levels it
touches in the multi-label summary of \cref{tab:methods}.

For every level, \cref{sec:feature,sec:prop,sec:topo,sec:learn,sec:fm} follow
the same template: (1) how the unified formulation is specialized; (2)
representative methods, described at the level of their update rules; (3)
differences from and connections to neighboring levels; and (4) strengths,
limitations, and computational cost.

\begin{table*}[t]
\centering
\caption{Multi-label summary of representative methods. \prim: primary
integration level; \pmark: secondary level. Feat = feature, Prop =
propagation, Topo = topology, Learn = learning, FM = foundation-model level.
Rec = recommendation; NC = node classification; LP = link prediction; Clust =
node clustering; GC = graph classification; Match = modality matching; MR =
modality retrieval; MA = modality alignment; G2T / G2I = graph-to-text /
graph-to-image generation. Tasks are those evaluated in each paper. Venues are
given as currently known; arXiv entries should be updated upon publication.}
\label{tab:methods}
\footnotesize
\setlength{\tabcolsep}{3pt}
\begin{tabular}{llcccccll}
\toprule
Method & Venue & Feat & Prop & Topo & Learn & FM & Graph-centric tasks & Modality-centric tasks \\
\midrule
MMGCN \citep{wei2019mmgcn}             & MM'19      &        & \prim  &        &        &        & Rec              & -- \\
MGAT \citep{tao2020mgat}               & IPM'20     &        & \prim  &        &        &        & Rec              & -- \\
GRCN \citep{wei2020graph}              & MM'20      &        & \pmark & \prim  &        &        & Rec              & -- \\
LATTICE \citep{zhang2021mining}        & MM'21      &        & \pmark & \prim  &        &        & Rec              & -- \\
FREEDOM \citep{zhou2023tale}           & MM'23      &        & \pmark & \prim  &        &        & Rec              & -- \\
BM3 \citep{zhou2023bm3}                & WWW'23     &        &        &        & \prim  &        & Rec              & -- \\
MMGL \citep{yoon2023mmgl}              & arXiv'23   &        &        &        &        & \prim  & --               & G2T \\
LGMRec \citep{guo2024lgmrec}           & AAAI'24    &        & \prim  & \pmark & \pmark &        & Rec              & -- \\
OMG-NAS \citep{cai2024omgnas}          & AAAI'24    &        & \prim  &        & \pmark &        & NC, GC           & -- \\
InstructG2I \citep{jin2024instructg2i} & NeurIPS'24 &        &        &        &        & \prim  & --               & G2I \\
UniGraph2 \citep{he2025unigraph2}      & WWW'25     &        & \pmark &        & \prim  &        & NC, LP, few-shot & G2T \\
MIG-GT \citep{hu2025miggt}             & AAAI'25    &        & \prim  &        &        &        & Rec              & -- \\
Graph4MM \citep{ning2025graph4mm}      & ICML'25    &        & \pmark &        &        & \prim  & NC (zero-shot)   & G2T \\
GraphGPT-O \citep{fang2025graphgpto}   & CVPR'25    &        &        &        &        & \prim  & --               & G2T, G2I \\
DMGC \citep{guo2025disentangling}               & MM'25      &        &        & \prim  & \pmark &        & Clust            & -- \\
MLaGA \citep{fan2026mlaga}             & KDD'26     &        &        &        & \pmark & \prim  & NC, LP           & -- \\
DGF \citep{zheng2026cross}               & KDD'26     &        & \prim  &        & \pmark &        & Clust            & -- \\
NTSFormer \citep{hu2026ntsformer}      & AAAI'26    &        & \prim  &        & \pmark &        & NC (cold-start)  & -- \\
LION \citep{li2026lion}                & arXiv'26   &        & \prim  &        &        &        & NC, LP, Clust    & MR, G2T, G2I \\
TMTE \citep{zhu2026tmte}               & arXiv'26      & \pmark &        & \prim  & \pmark &        & NC, LP, Clust    & MR, G2T, G2I \\
RoleMAG \citep{zuo2026rolemag}         & arXiv'26   &        & \prim  &        &        &        & NC, LP           & -- \\
CAMPA \citep{su2026campa}              & arXiv'26   &        & \prim  &        &        &        & NC, LP, Clust    & Match, MR, MA, G2I \\
CoMAG \citep{zhang2026context}           & arXiv'26    &        & \prim  & \prim  & \pmark &        & NC, LP, Clust    & Match, G2T, G2I \\
OptiMAG \citep{zuo2026optimag} & arXiv'26 & & & & \prim & & NC, LP, Clust & G2T, G2I \\
NSG-MoE \citep{zhang2026nsgmoe}        & arXiv'26   &        & \pmark & \prim  &        &        & NC, LP           & -- \\
ReCoG \citep{xue2026recog}             & arXiv'26   &        & \prim  & \prim  &        &        & NC, LP           & -- \\
GraphMNL \citep{wu2026multimodal}          & arXiv'26   &        &        &        & \prim  &        & NC               & -- \\
PLANET \citep{liu2026toward}           & arXiv'26    &        &        &        & \prim  &        & NC, LP, few-shot & G2T, G2I \\
SMGFM \citep{wu2026smgfm}              & arXiv'26    &        &        &        & \prim  &        & NC, LP, Clust    & MR \\
CHARM \citep{yang2026charm}            & arXiv'26   &        &        &        & \pmark & \prim  & NC, LP (zero-shot) & -- \\
FedGAMMA \citep{li2026toward}          & arXiv'26   &        &        &        & \prim  &        & NC, LP (federated) & Match, MR, MA, G2T \\
ICE \citep{li2026ice}                    & arXiv'26   &        &        &        & \prim  &        & NC, LP, few-shot & -- \\
Mario \citep{sun2026mario}             & CVPR'26    &        &        &        &        & \prim  & NC, LP, zero-shot & -- \\
\bottomrule
\end{tabular}
\end{table*}

\section{Feature Level: Encode, Fuse, then Propagate}
\label{sec:feature}

\textbf{Specialization of the unified formulation.}
Feature-level methods apply the fusion operator $F$ at $\ell=0$, so that a
single fused vector
\begin{equation}
\label{eq:feat-fuse}
\h_i^{(0)} = F\big(\{\mathbf{P}_m E_m(x_i^m;\theta_m)\}_{m\in\Mset_i}\big)
\end{equation}
is propagated by a standard GNN over the observed graph, i.e.,
$\Gamma_m(\A,\cdot)=\A$ for every $m$ and \cref{eq:unified-layer} reduces to a
single stream ($M=1$ after fusion). The design space therefore consists of the
encoders $E_m$, the fusion operator $F$, and the downstream GNN. Methods that
instead propagate each modality with its own GNN and fuse only at the output
(late fusion), such as MMGCN, are reviewed in \cref{sec:prop:decoupled}, because
their defining design choice concerns propagation rather than fusion.

\subsection{Modality Encoders}
\label{sec:feature:encoders}

We group encoders into three families.

\textbf{Unimodal encoders.}
Vision transformers such as ViT \citep{dosovitskiy2020image} and
self-supervised encoders such as DINOv2 \citep{oquab2023dinov2} encode images,
while language models such as T5 \citep{raffel2020t5} and RoBERTa
\citep{liu2019roberta} encode text. Their embedding spaces are not aligned, so
$\mathbf{P}_m$ must learn the alignment from the downstream task alone.

\textbf{Dual vision-language encoders.}
CLIP \citep{radford2021clip} and ImageBind \citep{girdhar2023imagebind}
map different modalities into a shared space through contrastive pretraining on
paired data. MM-GRAPH reports that aligned feature embeddings consistently
outperform unaligned ones across graph tasks \citep{zhu2025mmgraph};
accordingly, CLIP features are widely used in recent MAG work.

\textbf{MLLM-based encoders.}
Vision-language models such as LLaVA \citep{liu2023llava} and Qwen-VL
\citep{bai2023qwenvl}, as well as large language models such as LLaMA
\citep{touvron2023llama}, can produce either modality-specific embeddings or a
single joint embedding of the image and text of a node. MAGB finds that such
multimodal embeddings raise the performance ceiling of GNNs and alleviate the
imbalance between textual and visual attributes \citep{yan2025magb}. OpenMAG
further distinguishes static (frozen) from trainable encoding and supports 16
encoders in a unified framework \citep{wan2026openmag}.

\subsection{Fusion Operators Before Propagation}
\label{sec:feature:fusion}

Let $\mathbf{u}_i^m=\mathbf{P}_m E_m(x_i^m;\theta_m)\in\R^{d}$ denote the
projected embedding of modality $m$ of node $i$. Common choices of $F$ in
\cref{eq:feat-fuse} are:
\begin{itemize}
  \item \textbf{Concatenation:} with $\Vert$ denoting concatenation,
  $\h_i^{(0)}=\big\Vert_{m\in\Mset}\mathbf{u}_i^m$. This is the default in
  MM-GRAPH and MAGB baselines \citep{zhu2025mmgraph,yan2025magb}.
  \item \textbf{Sum or mean:} $\h_i^{(0)}=\frac{1}{|\Mset_i|}\sum_{m\in\Mset_i}\mathbf{u}_i^m$,
  which presupposes an aligned space (e.g., CLIP) and naturally handles missing
  modalities through $\Mset_i$.
  \item \textbf{Gated or attentive fusion:} with a learnable vector
  $\mathbf{q}\in\R^{d}$ and matrix $\mathbf{W}\in\R^{d\times d}$, the modality
  weights are
  $\beta_i^m=\operatorname{softmax}_m\!\big(\mathbf{q}^{\top}\tanh(\mathbf{W}\mathbf{u}_i^m)\big)$
  and $\h_i^{(0)}=\sum_{m\in\Mset_i}\beta_i^m\mathbf{u}_i^m$. The weights let the
  model down-weight a less informative modality per node.
  \item \textbf{Joint encoding:} a single multimodal encoder consumes the image
  and the text together and returns one embedding, so fusion happens inside the
  encoder.
\end{itemize}

The fused features are then propagated by a GNN such as GCN
\citep{kipf2017gcn}, GraphSAGE \citep{hamilton2017graphsage}, or GAT
\citep{velickovic2018gat}. For link prediction, MM-GRAPH additionally uses BUDDY
\citep{chamberlain2023buddy}, which augments message passing with subgraph
sketches.

\subsection{Analysis}
\label{sec:feature:analysis}

\textbf{Strengths.}
Feature-level methods are simple and efficient: encoder embeddings can be
precomputed once, after which training and inference cost the same as for an
ordinary single-stream GNN (see \emph{Complexity} below). They can also be
competitive on some datasets. For example, fused features with a single classical GNN such
as GraphSAGE perform comparably to models designed specifically for multimodal
graphs \citep{zhu2025mmgraph}.

\textbf{Limitations.}
After fusion, all modalities share one neighborhood and one propagation
operator. In terms of \cref{eq:unified-layer}, the model cannot assign
different neighbors, or different weights for the same neighbor, to different
modalities, and therefore cannot express modality-dependent neighbor utility
\citep{zuo2026rolemag}. Fusion also compresses modality-specific information
before the graph has a chance to contextualize it, which hurts
modality-centric tasks that need fine-grained correspondences
\citep{zhang2026context}. Moreover, the modalities of a node are not always
consistent: a product image may show a size chart rather than the item described
by its title \citep{xue2026recog}. Static fusion such as concatenation then
injects the mismatched modality as noise into every message that the node sends
and receives. Gated fusion (\cref{sec:feature:fusion}) can down-weight such a
modality, but only from the node's own features, without evidence from its
neighbors.

\textbf{Complexity.}
Let $C_m$ denote the cost of encoding one attribute of modality $m$. Encoding
costs $O\big(N\sum_m C_m\big)$ once if encoders are frozen (and per epoch if
they are trained), and an $L$-layer GNN costs $O\big(L(|\E|d+Nd^2)\big)$ per
epoch. In practice the encoder term dominates whenever encoders are fine-tuned,
which is the main reason why most MAG benchmarks default to frozen features.

\section{Propagation Level: Modality-Aware Message Passing}
\label{sec:prop}

\textbf{Specialization of the unified formulation.}
Propagation-level methods keep a separate representation $\h_i^{m,(\ell)}$ for
each modality and design the message functions $\psi_{m\leftarrow m'}$ and the
coupling coefficients $\alpha_{ij}^{m\leftarrow m'}$ in
\cref{eq:unified-msg}. The key distinction is whether the off-diagonal
coefficients ($m'\neq m$) are zero (\emph{modality-decoupled}, \cref{sec:prop:decoupled})
or not (\emph{cross-modal coupled}, \cref{sec:prop:coupled}).

\subsection{Modality-Decoupled Propagation with Late Fusion}
\label{sec:prop:decoupled}

Modality-decoupled methods set $\alpha_{ij}^{m\leftarrow m'}=0$ for
$m'\neq m$, so \cref{eq:unified-msg,eq:unified-layer} become $M$ independent GNNs whose outputs
meet only in $F$:
\begin{equation}
\label{eq:decoupled}
\begin{aligned}
\mathbf{c}_{ij}^{m,(\ell)} &= \alpha_{ij}^{m\leftarrow m}\,
\psi_{m\leftarrow m}\big(\h_j^{m,(\ell)}\big),\\
\h_i^{m,(\ell+1)} &= \UPD_m\Big(\h_i^{m,(\ell)},\,
\AGG_{j\in\Nb^{m}(i)}\mathbf{c}_{ij}^{m,(\ell)}\Big),\\
\z_i &= F\big(\{\h_i^{m,(L)}\}_{m\in\Mset}\big).
\end{aligned}
\end{equation}

\textbf{MMGCN} \citep{wei2019mmgcn} instantiates \cref{eq:decoupled} on a
user--item bipartite graph: each modality runs its own graph convolution over
the interaction graph, so that user preferences are modeled separately per
modality, and the modality-specific outputs are combined at the end. In the
recommendation setting it also maintains ID embeddings that are shared across
modalities.

\textbf{MGAT} \citep{tao2020mgat} replaces uniform aggregation within each
modality-specific graph by a gated attention mechanism, so that
$\alpha_{ij}^{m\leftarrow m}$ becomes a learned, modality-specific weight of
neighbor $j$, and then combines the modality-specific representations.

\textbf{LGMRec} \citep{guo2024lgmrec} combines collaborative and modality-specific local graph embeddings with modality-aware global hypergraph learning to capture dependencies beyond local neighborhoods.

\textbf{MIG-GT} \citep{hu2025miggt} assigns each modality its own
receptive field, using separate GNNs per modality whose depth can differ, and
adds a sampling-based global transformer. In terms of
\cref{eq:decoupled}, each modality has its own number of layers $L_m$, which
acknowledges that the optimal propagation depth is modality-dependent.

\textbf{Discussion.}
Modality-decoupled propagation preserves modality-specific signals and accommodates distinct neighborhoods and propagation depths. However, under independent modality-specific message passing, it remains cross-modally insensitive before fusion (\cref{obs:jacobian}): a neighbor's visual features cannot influence the target's text-channel aggregation. Its propagation cost scales approximately linearly with the number of modalities $M$, assuming comparable per-modality complexity.


\subsection{Cross-Modal Coupled Propagation}
\label{sec:prop:coupled}

Coupled methods allow $\alpha_{ij}^{m\leftarrow m'}\neq0$ for $m'\neq m$, so
that every layer exchanges information across modalities \emph{and} across
nodes. Recent methods differ mainly in how they decide which neighbor should
send which modality to which stream.

\textbf{ReCoG} \citep{xue2026recog} starts from the limitation stated in
\cref{obs:jacobian}: in a modality-decoupled model, the image of a neighbor can
never affect how the text of the target node is aggregated. ReCoG proves that
such layerwise cross-modal information flow is impossible in decoupled and
late-fusion architectures regardless of their capacity, and accordingly keeps a
text stream and a vision stream that exchange messages at every layer, both
within a node and across neighbors. To state the update, let $\tilde{\A}$ be the
refined adjacency produced by its structure refiner (\cref{sec:topo:refine}),
symmetrically normalized with self-loops, let $\mathbf{W}^{(\ell)}\in\R^{d\times
d}$ be the weight matrix of layer $\ell$, shared by both streams, let $\sigma$
be a nonlinearity, and let $\gamma,\rho\in(0,1)$ be learnable coupling
strengths. The text stream is updated as
\begin{equation}
\label{eq:recog}
\begin{aligned}
\h_i^{\mathrm{t},(\ell+1)} = \sigma\Big(\mathbf{W}^{(\ell)}\Big(
&\textstyle\sum_{j}\tilde{\A}_{ij}\,\h_j^{\mathrm{t},(\ell)}
+\gamma\,\h_i^{\mathrm{v},(\ell)}\\
&+\rho\textstyle\sum_{j}\tilde{\A}_{ij}\,\h_j^{\mathrm{v},(\ell)}\Big)\Big),
\end{aligned}
\end{equation}
where the sums run over the neighbors $j$ of $i$ in the refined graph; the
vision stream is updated symmetrically, with $\mathrm{t}$ and $\mathrm{v}$
exchanged. The three terms are intra-modal neighbor aggregation, cross-modal
self-injection, and cross-modal neighbor aggregation. In terms of
\cref{eq:unified-msg}, $\alpha_{ij}^{m\leftarrow m}=\tilde{\A}_{ij}$ and
$\alpha_{ij}^{m\leftarrow m'}=\rho\tilde{\A}_{ij}$ for $m'\neq m$, with an
additional cross-modal self-loop of weight $\gamma$. The last term is what
modality-decoupled models lack: for example, it lets the images of co-purchased
products sharpen the text representation of a product whose title is
uninformative.

Stacking the two streams as
$\Hm^{(\ell)}=[\Hm^{\mathrm{t},(\ell)};\Hm^{\mathrm{v},(\ell)}]\in\R^{2N\times d}$
shows that \cref{eq:recog} is an ordinary GNN layer on a multiplex graph with
$2N$ nodes, a text copy and a vision copy of every node, whose adjacency is
\begin{equation}
\label{eq:recog-mux}
\tilde{\A}_{\mathrm{mux}}=
\begin{bmatrix}
\tilde{\A} & \gamma\mathbf{I}+\rho\tilde{\A}\\
\gamma\mathbf{I}+\rho\tilde{\A} & \tilde{\A}
\end{bmatrix}\in\R^{2N\times 2N}.
\end{equation}
This view makes the role of the coupling strengths explicit. If
$\mu_1\ge\dots\ge\mu_N$ are the eigenvalues of $\tilde{\A}$, the eigenvalues of
$\tilde{\A}_{\mathrm{mux}}$ are $(1+\rho)\mu_i+\gamma$ for modes in which the
two modalities agree and $(1-\rho)\mu_i-\gamma$ for modes in which they
oppose, so $\gamma$ and $\rho$ act as a learned spectral filter that amplifies
agreement between modalities. Because $\mathbf{W}^{(\ell)}$ is shared, each
layer also scales the text--vision discrepancy
$\Hm^{\mathrm{t},(\ell)}-\Hm^{\mathrm{v},(\ell)}$ by at most
$\|\mathbf{W}^{(\ell)}\|\big((1-\rho)\|\tilde{\A}\|+\gamma\big)$, so the two
streams contract toward a shared space when this factor is below one. The
refiner and the GNN are trained jointly (\cref{sec:topo:learned}).

\textbf{RoleMAG} \citep{zuo2026rolemag} makes the coupling neighbor-specific. It starts from the observation that the same neighbor need not be equally useful for all modalities, assigns each edge soft weights $\pi^{\mathrm{sh}}_{ij}$, $\pi^{\mathrm{c}}_{ij}$, and $\pi^{\mathrm{het}}_{ij}$ (summing to one) over three roles, \emph{shared}, \emph{complementary}, and \emph{heterophilous}, and routes each role through its own propagation channel. Let $\mathcal{N}_{m'\to m}(i)$ be the $K$ neighbors of $i$ with the largest complementary weights for completing modality $m$ from modality $m'$, and $\psi_{\mathrm{sh}}$, $\psi_{m\leftarrow m'}$, and $\psi_{\mathrm{het}}$ the message functions of the three channels. Schematically,
\begin{equation}
\begin{aligned}
\mathbf{h}_i^m \leftarrow \mathrm{UPD}_m\Big(\mathbf{h}_i^m,\;
&\mathrm{AGG}_{j\in\mathcal{N}(i)}\,\pi^{\mathrm{sh}}_{ij}\,\psi_{\mathrm{sh}}(\mathbf{h}_j^m),\\
&\mathrm{AGG}_{j\in\mathcal{N}_{m'\to m}(i)}\,\pi^{\mathrm{c}}_{ij}\,\psi_{m\leftarrow m'}(\mathbf{h}_j^{m'},\mathbf{h}_i^m),\\
&\mathrm{AGG}_{j\in\mathcal{N}(i)}\,\pi^{\mathrm{het}}_{ij}\,\psi_{\mathrm{het}}(\mathbf{h}_j^m)\Big).
\end{aligned}
\end{equation}
The first channel performs shared smoothing. The second performs cross-modal completion, with complementary neighbors selected separately for each direction (top-$K$ for text$\to$image and for image$\to$text). The third separates heterophilous evidence from shared smoothing. The design rests on two
empirical findings: neighbor utility is modality-dependent, and complementary
completion is directional, depending on which modality is currently weaker.

\textbf{CoMAG} \citep{zhang2026context} uses a similar propagation: the modality-$m$ stream aggregates its own modality and the
mean of the other modalities over a context graph $\mathbf{Q}_\tau$ for task
$\tau$. It differs in that propagation is linear and multi-hop, the coupling
coefficients are conditioned on the task, and an initial residual to the
projected input makes the iteration contractive and mitigates over-smoothing.
Its main contributions lie at the other levels: the reliability-aware context
graph (\cref{sec:topo:refine}) and hop-token alignment with a
shared--private decomposition (\cref{sec:learn:align}).

\textbf{CAMPA} \citep{su2026campa} and \textbf{LION} \citep{li2026lion} couple
modalities inside a multi-hop propagation that is computed once before
training, so that only the subsequent aggregation is learned; their efficiency
is discussed in \cref{sec:cross:efficiency}. CAMPA observes that diffusing each
modality separately makes text and image features drift apart as the number of
hops grows. It therefore adds to every hop a cross-modal neighbor term, as in
\cref{eq:recog} but with fixed coefficients, over edges reweighted by
cross-modal similarity, and aligns the hop features of the two modalities with
attention before fusing them. LION represents the text and image features of a
node as a single multivector in a Clifford algebra, so that propagating these
multivectors mixes the two modalities, and then learns how to weight the
resulting components and propagation depths.

\subsection{Analysis: Differences, Connections, and Cost}
\label{sec:prop:analysis}

\textbf{What separates the two families.}
\cref{obs:jacobian} gives a precise criterion: modality-decoupled models are
cross-modally insensitive before fusion, whereas coupled models are not. The
difference matters exactly when cross-node, cross-modal evidence is
informative, for example when a node's title is uninformative but its
neighbors' images are. Coupled methods, in turn, differ in how they control
where cross-modal messages flow. ReCoG and CoMAG apply one coupling strength to
all edges and control the flow through the graph itself, which ReCoG learns end
to end with the representations and CoMAG builds from edge reliability. RoleMAG
instead decides, per neighbor and per direction, whether cross-modal completion
should happen at all.

\begin{table*}[t]
\centering
\caption{Approximate cost of propagation-level designs, excluding encoders and
structure learning. \emph{Offline}: computed once before training. $L$: layers;
$K$: hops; $M$: modalities; $N$: nodes; $|\E|$: edges; $d$: hidden size.}
\label{tab:prop-complexity}
\small
\begin{tabular}{llll}
\toprule
Design & Examples & Offline & Per epoch \\
\midrule
Early fusion (\cref{sec:feature})              & GCN, SAGE on fused features & --             & $O\big(L(|\E|d+Nd^2)\big)$ \\
Modality-decoupled (\cref{sec:prop:decoupled}) & MMGCN, MGAT                 & --             & $O\big(LM(|\E|d+Nd^2)\big)$ \\
Coupled                                        & ReCoG                       & --             & $O\big(LM(|\E|d+Nd^2)\big)$ \\
Precomputed coupled + hop attention            & CAMPA                       & $O(KM^2|\E|d)$ & $O\big(NMKd^2+NM^2K^2d\big)$ \\
\bottomrule
\end{tabular}
\end{table*}

\textbf{Complexity.}
\cref{tab:prop-complexity} gives orders of magnitude. Cross-modal coupling as in
ReCoG costs the same order as modality-decoupled propagation: because
aggregation is linear, the intra-modal and cross-modal neighbor terms of
\cref{eq:recog} can be merged into a single aggregation per stream, so each
layer needs two aggregations, as many as a modality-decoupled model. CAMPA computes its propagation before training, so its
per-epoch cost does not depend on $|\E|$, but the propagation is then fixed and
cannot adapt to the task.

\section{Topology Level: Multimodal Semantics Reshape Structure}
\label{sec:topo}

\textbf{Specialization of the unified formulation.}
Topology-level methods design the structure operator $\Gamma_m$ in
\cref{eq:unified-structure}. The observed graph $\A$ is often noisy,
incomplete, or task-agnostic (e.g., co-purchase edges that reflect
co-occurrence rather than the target task). In MAGs, the usefulness of an edge
also depends on the modality: two adjacent nodes may have similar text but
dissimilar images, so an edge that is homophilous for one modality can be
heterophilous for another \citep{zuo2026rolemag,guo2025disentangling}. Multimodal
semantics can therefore be used to build, refine, or learn the graph on which
propagation runs, and to handle edges whose usefulness differs across
modalities.

\subsection{Modality-Specific Graph Construction}
\label{sec:topo:construction}

The simplest structure operator builds a $k$-nearest-neighbor graph in each
modality space. With $s(\cdot,\cdot)$ a similarity function (typically cosine)
and $\operatorname{TopK}_i$ the $k$ most similar nodes to $i$,
\begin{equation}
\label{eq:knn}
\tilde{\A}^{m}_{ij} = s(\x_i^m,\x_j^m)\cdot\mathbb{I}\big[j\in\operatorname{TopK}_i\big(s(\x_i^m,\cdot)\big)\big].
\end{equation}
This idea originated in multimodal recommendation. LATTICE
\citep{zhang2021mining} learns modality-aware item--item graphs from
multimodal features, combines them into a latent item graph, and propagates on
it. FREEDOM \citep{zhou2023tale} shows that the item--item graph
can be frozen after construction, and additionally denoises the user--item graph
by degree-sensitive edge pruning. These methods are mostly evaluated on
recommendation; we include them because constructed modality graphs are a
recurring building block of later MAG methods.

\subsection{Refining the Observed Graph}
\label{sec:topo:refine}

Instead of building a graph from scratch, a refiner keeps the observed graph and
corrects it: it re-scores observed edges to suppress spurious ones and adds
semantic edges that the observed graph misses. In multimodal recommendation,
GRCN \citep{wei2020graph} re-scores the observed user--item edges: for
each modality, it represents a user's preference as a prototype refined by
neighbor routing, treats the affinity between an item's content and this
prototype as the confidence that the interaction is a true positive, and prunes
edges softly according to these affinities before graph convolution. GRCN only
down-weights observed edges; ReCoG \citep{xue2026recog} additionally recovers missing edges. It refines the observed adjacency using multimodal semantic evidence through two complementary refiners: a similarity-based refiner and a learned residual refiner. For scalable mini-batch training, both restrict candidate pairs to a target node and a node in its sampled subgraph, covering observed edges for reweighting and candidate non-edges for recovery. The residual refiner further employs coarse-to-fine candidate selection to reduce pairwise scoring costs.
Specifically, let $\mathbf{A}_{ij}\in\{0,1\}$ denote the observed adjacency, $\mathbf x_i^{\mathrm t}$ and $\mathbf x_i^{\mathrm v}$ the text and visual encoder features, and $\operatorname{Clamp}$ clipping to $[0,1]$. The cosine refiner computes a weighted multimodal similarity
\begin{align}
s_{ij}=\max\big(0,\lambda\cos(\mathbf x_i^{\mathrm v},\mathbf x_j^{\mathrm v})+(1-\lambda)\cos(\mathbf x_i^{\mathrm t},\mathbf x_j^{\mathrm t})\big),
\end{align}
and combines the observed topology with the top-$k$ semantic connections:
\begin{align}
\tilde{\mathbf{A}}_{ij} = \mathrm{Clamp}\big(\alpha\mathbf{A}_{ij} + (1-\alpha)\,s_{ij}\,\mathbb{I}[j\in\mathcal{N}_{\mathrm{sem}}(i)]\big),
\end{align}
where $\mathcal N_{\mathrm{sem}}(i)$ contains the $k$ highest-scoring candidates of node $i$, and $\lambda,\alpha\in(0,1)$ are learnable scalars. This allows low-similarity observed edges to be down-weighted and similar non-edges to be added, but limits corrections to cosine-based similarity.

The residual refiner instead learns signed, pair-specific corrections:
\begin{align}
\tilde{\mathbf{A}}_{ij} = \mathrm{Clamp}\big(\mathbf{A}_{ij} + \delta\tanh(f_\theta(\boldsymbol{\phi}_{ij}))\big),
\end{align}
where $f_\theta$ is a learnable MLP, $\delta\in(0,1)$ controls the correction magnitude, and $\boldsymbol\phi_{ij}$ combines endpoint features, element-wise products, and absolute differences across modalities. Negative corrections suppress observed edges, while positive corrections can introduce missing connections. To reduce scoring costs, the residual refiner first selects candidates using low-dimensional similarity before applying the MLP. Compared with fixed cosine-based scoring, it enables more expressive, task-adaptive topology refinement.

\subsection{Learned and Co-Evolving Topology}
\label{sec:topo:learned}

Graph structure learning (GSL) on unimodal graphs learns the graph jointly with
the GNN. IDGL \citep{chen2020idgl} iteratively refines a metric-learned graph
together with node embeddings; Pro-GNN \citep{jin2020graph} learns a clean
graph under sparsity, low-rank, and feature-smoothness priors to defend against
adversarial edges; SLAPS \citep{fatemi2021slaps} introduces self-supervised feature denoising to improve structure learning; and NodeFormer \citep{wu2022nodeformer}
scales all-pair message passing with a kernelized Gumbel--Softmax operator,
learning latent connections without an explicit graph. These methods learn the
graph in a single representation space. Applied to MAGs, they score pairs from
concatenated features, so modality-specific evidence is collapsed before
refinement and the modalities do not exchange information during propagation
\citep{xue2026recog}. Multimodal GSL must additionally decide which modalities
induce the learned structure and how that structure feeds back into each
modality.

\textbf{ReCoG} \citep{xue2026recog} jointly optimizes the multimodal graph refiners introduced in \cref{sec:topo:refine} with coupled cross-modal propagation (\cref{eq:recog}), allowing graph structure and multimodal representations to mutually reinforce each other. Topology refinement suppresses noisy edges and recovers semantically meaningful connections, providing a more reliable structure for cross-modal information exchange. Conversely, coupled propagation enables complementary information to flow across modalities and neighboring nodes, while its task gradients guide the refiner toward connections that benefit joint representation learning. Without refinement, cross-modal propagation remains constrained by noisy or incomplete topology; without coupling, the refined structure cannot fully exploit cross-modal neighbor interactions. Ablation results confirm their complementarity: removing either component degrades performance, while removing both yields the largest degradation, highlighting the importance of jointly learning graph structure and multimodal representations.

\textbf{TMTE} \citep{zhu2026tmte}, a concurrent work, adopts a different co-evolution mechanism. It learns task-adaptive topology through multi-perspective similarity metrics over modality-specific and fused embeddings, using a node--anchor approximation for scalability. The evolved graph guides smoothness-regularized learning of fused representations, while cross-modal contrastive alignment preserves consistency across modalities. Updated representations then guide subsequent topology refinement. Unlike ReCoG's explicit cross-modal neighbor exchange between separate modality streams, TMTE performs graph propagation on fused representations and learns non-negative similarity-based topology mixed with the observed graph.

\subsection{Modality-Dependent Heterophily}
\label{sec:topo:hetero}

Several methods address modality-dependent neighborhood inconsistencies, where connected nodes may be similar in one modality but dissimilar in another.

OptiMAG \citep{zuo2026optimag} addresses the same inconsistency without rewiring the graph, through a learning-level regularizer (Section~\ref{sec:learn:align}).

\textbf{DMGC} \citep{guo2025disentangling} explicitly disentangles homophilous and heterophilous relations by constructing a cross-modal consensus graph and modality-specific heterophily-aware graphs, applying low-pass and high-pass filters to preserve shared semantic consistency and modality-specific distinctions for clustering.

\textbf{NSG-MoE} \citep{zhang2026nsgmoe} addresses the related problem of
modality confusion, in which a general-purpose GNN mixes messages from different
modalities indiscriminately. It splits each node into modality-specific
components, rewires the graph into within- and cross-modality relations, and
assigns relation-aware experts to the different message flows.

Together with role-aware routing at the propagation level (\cref{sec:prop:coupled}) and modality-guided topology refinement (\cref{sec:topo:refine}), these methods offer complementary strategies for handling inconsistent multimodal neighborhoods. Importantly, modality disagreement does not necessarily imply uninformative edges: such relations may contain useful complementary or heterophilous signals rather than noise.

\subsection{Analysis}
\label{sec:topo:analysis}

\textbf{Reweighting versus removing edges.}
The methods of this section differ in which corrections to the observed graph
they can express:
\begin{itemize}
  \item \emph{Additive mixtures}, $\tilde{\A}=\alpha\A+(1-\alpha)\A^{\mathrm{sem}}$
  with $\A^{\mathrm{sem}}\ge0$, such as the cosine refiner of ReCoG and the
  topology of TMTE, add semantic connections and reweight observed edges, but
  keep every observed edge with a positive weight when $\alpha>0$.
  \item \emph{Multiplicative gating}, $\tilde{\A}=\A\odot\mathbf{R}$ with
  $\mathbf{R}\in[0,1]^{N\times N}$, such as the soft pruning of GRCN, suppresses
  unreliable observed edges but introduces no new ones; exact removal requires
  zero-valued gates or explicit thresholding.
  \item \emph{Signed residual corrections},
  $\tilde{\A}=\operatorname{Clamp}(\A+\boldsymbol{\Delta})$, such as the
  residual refiner of ReCoG, can both add or strengthen edges and weaken
  observed ones through pair-specific adjustments. Exact removal requires a
  correction that reaches $-\A_{ij}$; with ReCoG's bounded correction
  ($\delta<1$), observed edges are weakened but remain positive.
  \item \emph{Hard selection}, such as the top-$k$ selection in \cref{eq:knn} or
  of semantic candidates in ReCoG, sparsifies candidate connections through a
  discrete, non-differentiable choice; whether observed edges are removed depends
  on how the selected graph is combined with the original one.
\end{itemize}
Operators that only add or only down-weight edges suit datasets dominated by
missing or by spurious relations, respectively, whereas signed corrections
address both. Because disagreement between modalities does not necessarily
mark an edge as noise (\cref{sec:topo:hetero}), operators that weaken edges
without removing them also retain heterophilous connections that may still be
informative.

\textbf{One shared graph or modality-specific structure.}
Methods also differ in how many structures they maintain. ReCoG and TMTE refine a
single graph shared by all modalities, which keeps the cost of one graph and, in
ReCoG, lets both streams exchange cross-modal messages along the same edges, but
assigns each edge one weight for all modalities. DMGC builds modality-specific
heterophily-aware graphs alongside a consensus graph, and NSG-MoE splits each
node into modality-specific components, so that an edge can be treated
differently in each modality, at the cost of several graphs or a larger
heterogeneous graph. OptiMAG keeps the observed graph and instead encourages
consistency between modality-specific structures and the topology through its
training objective.

\textbf{Topology as an implicit cross-modal channel.}
Cross-modal interaction can arise not only from coupled propagation but also
from the graph itself. When the structure operator $\Gamma_m$ is computed from
other modalities, as in the cosine refiner of ReCoG, whose scores mix text and
vision similarities, or in $k$-NN over fused features, condition (ii) of
\cref{obs:jacobian} no longer holds: a change in one modality can change the
neighbors, or the neighbor weights, through which another modality is
aggregated, even when the feature streams propagate separately. This holds
whether the graph is precomputed or learned; learning only makes the channel
adaptive. Ablations of cross-modal propagation should therefore control for the
multimodal information already contained in the graph. ReCoG reports such a control: on Ele-fashion with CLIP features and a GCN backbone, its refined graph already improves a single-stream GCN, and coupled propagation on the same refined graph improves it further \citep{xue2026recog}.

\textbf{Cost.}
Explicit all-pair structure learning requires $O(N^2d)$ computation and
$O(N^2)$ memory when the full affinity matrix is materialized, which limits
scalability. Practical alternatives include node--anchor approximations
\citep{zhu2026tmte} and target-centric candidate selection within sampled
subgraphs, with coarse-to-fine filtering before expensive pairwise scoring
\citep{xue2026recog}.

\section{Learning Level: Graph-Aware Alignment and Pretraining}
\label{sec:learn}

\textbf{Specialization of the unified formulation.}
Learning-level integration is defined by the training objective $\mathcal{L}$
and the pretraining protocol, even when methods also introduce new feature
fusion or message-passing operators. We organize existing methods into three directions:
(i) graph-aware alignment and regularization objectives, (ii) self-supervised
and imbalance-aware learning, and (iii) cross-graph pretraining for
transferable multimodal representations. For foundation models, this section
focuses on learning objectives and transfer protocols, while their
architectural interfaces with large models are discussed in \cref{sec:fm}.

\subsection{Graph-Aware Alignment and Regularization}
\label{sec:learn:align}

Cross-modal alignment encourages consistent representations across modalities,
but excessive alignment may suppress modality-specific information. Graph-aware
objectives address this trade-off by incorporating structural context into
alignment, regularizing consistency with the graph, or explicitly separating
shared and private information.

\textbf{Cross-modal contrastive alignment.}
A common objective aligns different modalities of the same node while
contrasting them against other nodes. Let $\mathbf{e}_i^m\in\mathbb{R}^d$ denote the
representation of node $i$ in modality $m$, $s(\cdot,\cdot)$ cosine similarity,
$t>0$ a temperature, and $\mathcal{B}$ a mini-batch. A standard objective is
\begin{equation}
\label{eq:infonce}
\mathcal{L}_{\mathrm{align}}
=-\sum_{i\in\mathcal{B}}\sum_{m\neq m'}
\log\frac{\exp\big(s(\mathbf{e}_i^m,\mathbf{e}_i^{m'})/t\big)}
{\sum_{j\in\mathcal{B}}\exp\big(s(\mathbf{e}_i^m,\mathbf{e}_j^{m'})/t\big)}.
\end{equation}
Applied to raw encoder features, this is CLIP-style alignment without graph
context. Graph-aware methods instead align representations that incorporate
structural information: LGMRec \citep{guo2024lgmrec} contrasts the visual and
textual global hypergraph embeddings of the same user or item, TMTE
\citep{zhu2026tmte} aligns each modality embedding with the other modalities
and with a fused representation smoothed over the evolving topology, and CoMAG
\citep{zhang2026context} uses a matching objective on graph-contextualized
modality representations to distinguish matched from mismatched pairs.

\textbf{Structural consistency regularization.}
Objectives can also encourage modality representations to respect relational
structure rather than only matching one another. OptiMAG
\citep{zuo2026optimag} introduces a plug-in unbalanced fused
Gromov--Wasserstein regularizer that aligns modality-specific semantic
geometry with graph distances derived from personalized PageRank. Its relaxed
transport constraints reduce the influence of structurally inconsistent
correspondences (see also \cref{sec:topo:hetero}). This approach injects structural consistency through the training loss and
can complement existing propagation operators.

\textbf{Shared--private regularization.}
Forcing complete agreement may remove information unique to individual
modalities. One approach decomposes each node representation into a shared
consensus $\mathbf{s}_i$ and modality-private residuals
$\mathbf{p}_i^{(m)}$, then discourages their redundancy. CoMAG
\citep{zhang2026context}, for example, employs squared cosine similarity:
\begin{equation}
\label{eq:orth}
\mathcal{L}_{\perp}
=\frac{1}{NM}\sum_{i=1}^{N}\sum_{m=1}^{M}
\bigg(\frac{\mathbf{s}_i^{\top}\mathbf{p}_i^{(m)}}
{\|\mathbf{s}_i\|_2\,\|\mathbf{p}_i^{(m)}\|_2}\bigg)^{2}.
\end{equation}
Graph-centric tasks can use the consensus, while modality-centric tasks
also incorporate private components. A related role-separation objective
appears in SMGFM's frequency-aware pretraining (\cref{sec:learn:mgfm}).

\textbf{Contrastive clustering.} DGF \citep{zheng2026cross} targets node clustering on MAGs. It combines a node-domain graph filter with a feature-domain filter that denoises across feature dimensions, and trains with a tri-cross contrastive objective at the levels of modalities, neighborhoods, and communities.

\subsection{Self-Supervised and Imbalance-Aware Objectives}
\label{sec:learn:ssl}

Beyond the task loss, training can exploit multimodal attributes and graph
relations to generate self-supervision, or use branch reliability to moderate
cross-modal guidance. The latter may still depend on supervised labels and
should not be conflated with label-free self-supervised learning.

\textbf{Bootstrapped and graph-guided self-supervision.}
In multimodal recommendation, BM3 \citep{zhou2023bm3} avoids negative samples
and graph augmentation: it generates latent views by dropout, treats the
dropped view as a stop-gradient target, and jointly optimizes interaction
reconstruction with intra- and inter-modality alignment. On MAGs, MLaGA
\citep{fan2026mlaga} pretrains a structure-aware multimodal aligner with a
graph contrastive objective that pulls a node's fused text--image
representation toward its sampled neighbors, before the aligned representation
is passed to an LLM (\cref{sec:fm}).

\textbf{Reliability-aware negative learning.}
A distinct challenge is \emph{node-level branch imbalance}: structural,
textual, and visual evidence may differ in reliability across nodes, and
conventional cross-branch alignment can propagate incorrect supervision when
the dominant branch is unreliable. GraphMNL \citep{wu2026multimodal} scores the
reliability of each branch per node from prediction confidence, neighborhood
consistency, and uncertainty. Instead of requiring weaker branches to imitate
the dominant prediction, it uses reliability-controlled negative learning that
suppresses unlikely non-target classes while retaining ordinary target
supervision.

\subsection{Multimodal Graph Foundation Models}
\label{sec:learn:mgfm}

Multimodal graph foundation models (MGFMs) aim to learn representations or
prediction mechanisms that are reusable across graph domains and tasks. From
the learning-level perspective, they are defined by their cross-graph
pretraining objectives and transfer protocols rather than by architectural
scale alone.

\textbf{Reconstruction objectives.}
A common strategy learns transferable representations by reconstructing
masked attributes or structural relationships. UniGraph2
\citep{he2025unigraph2} encodes nodes with frozen modality-specific encoders,
aligns features across modalities and domains through a mixture-of-experts
module, and reconstructs masked features with domain-specific decoders and
graph structure with a shared shortest-path-distance decoder. PLANET \citep{liu2026toward} reconstructs each masked modality from its own and
from complementary modalities together with observed edges, whereas SMGFM
\citep{wu2026smgfm} combines masked feature and edge reconstruction with
frequency-specific learning objectives.

\textbf{Alignment that preserves modality-specific evidence.}
Rather than aligning all modalities uniformly, recent MGFMs decide which
components to align. PLANET \citep{liu2026toward} separates local interaction
from global alignment: embedding-wise domain gating incorporates complementary
neighbor modalities, and node-wise discretization maps modality embeddings
into a shared discretized semantic space with text-anchored alignment and
quantization objectives. SMGFM \citep{wu2026smgfm} decomposes modality-specific
signals into graph-frequency bands with Chebyshev filters, aligns
low-frequency consensus components, and preserves high-frequency
modality-specific evidence with energy- and direction-preserving losses.

\textbf{Transfer protocols and settings.}
MGFMs also differ in how and where they transfer. CHARM \citep{yang2026charm}
targets zero-shot transfer to unseen graphs without target-domain parameter
updates, encoding retrieved contexts built from hierarchical semantic anchors
and modality-complementary neighbors as graph tokens for a frozen LLM
(\cref{sec:fm:softtokens}). FedGAMMA \citep{li2026toward} extends cross-graph learning to federated
clients, combining shared--private semantic enhancement and
optimal-transport-based alignment with topology-aware fusion and
client-affinity aggregation, followed by prompt-based adaptation. 

\subsection{Discussion}
\label{sec:learn:discussion}

Learning-level methods differ mainly in the supervision they add to the task
loss. Alignment and regularization objectives constrain agreement between
modalities and with the graph; self-supervised and imbalance-aware objectives
construct signals from augmented views, structural relationships, or branch
reliability; and pretraining objectives aim at representations that transfer
across graphs. A question recurs across all three: which information should be
aligned and which modality-specific differences should be preserved.
Contrastive objectives push toward agreement, whereas shared--private
decompositions, frequency-specific pretraining, and reliability-aware
supervision deliberately keep parts of each modality apart, because cross-modal
agreement is not always reliable supervision when modality features or graph
context are noisy. Explicit learning objectives should be distinguished from architectural
alignment through attention or coupled propagation
(\cref{sec:prop:coupled}), even though both may be optimized jointly.
Cross-graph pretraining additionally changes the training data and transfer
protocol rather than merely adding a loss term. For pretrained
models, evaluations should further separate in-domain performance,
cross-domain adaptation, and genuine zero-shot transfer, and account for the
scope and cost of pretraining.
\section{Foundation-Model Level: Bringing Graphs into (M)LLMs}
\label{sec:fm}

\textbf{Specialization of the unified formulation.}
Foundation-model-level integration connects multimodal graph structure to
pretrained language, vision--language, or generative backbones. In
\cref{eq:unified-fm}, an interface $\mathcal{T}$ converts an ego-subgraph
$\mathcal{G}_i$ and its multimodal attributes into inputs to a foundation
model $\Phi$, or modifies its computation using graph structure. The
principal choices concern (i) how structural information is represented,
(ii) whether graph reasoning occurs in a graph-side encoder or within the
foundation model, and (iii) how the pretrained backbone is adapted. The tasks also differ: some methods perform graph-centric prediction, mainly
node classification and link prediction, whereas others generate the text
(G2Text), the image (G2Image), or both for a target node. We first
review serialization and graph-aware interfaces, then direct MLLM prediction,
and finally graph-conditioned generation. The first three concern model
interfaces and prediction; the last concerns a complementary output setting,
so these categories are not mutually exclusive.

\subsection{Graph Serialization and Prompting}
\label{sec:fm:sequence}

The most direct interface represents graph context as a sequence of node
attributes, structural descriptors, and prompts that a pretrained model can
consume. Its appeal is compatibility with existing language and multimodal
models, but serializing a graph does not by itself preserve all relations
among the included nodes.

\textbf{MMGL} \citep{yoon2023mmgl} targets G2Text, namely Wikipedia section
summarization, and studies how pretrained language models can use multiple
multimodal neighbors beyond one-to-one image--caption pairs. It compares ways to encode neighbors (raw text versus
frozen embeddings, self- versus cross-attention), ways to encode the structure
among them (sequential or Laplacian positional encodings, GNN embeddings), and
parameter-efficient adaptation (prefix tuning, LoRA, Flamingo-style
cross-attention). MMGL is thus not restricted to plain attribute
concatenation: the central question is how much relational information the
language-model interface can retain.

\textbf{GraphGPT-O} \citep{fang2025graphgpto} targets joint G2Text and G2Image,
generating both the text and the image of a target node. It compares ways to
linearize the multimodal neighborhood into an MLLM input, varying which modalities are
included, their order, and the number of neighbors. Its results motivate a
hierarchical aligner (\cref{sec:fm:softtokens}), and it further compares
sequential and parallel generation of interdependent text and images. Both
works illustrate the trade-off between the flexibility of prompting and the
difficulty of representing irregular neighborhoods in a finite token budget.

\subsection{Graph-Aware Token and Attention Interfaces}
\label{sec:fm:softtokens}

Rather than serializing all neighbor content, graph-aware interfaces
summarize structural context in continuous tokens or incorporate graph
relations directly into a foundation model's attention.

\textbf{Graph-to-token interfaces.}
MLaGA \citep{fan2026mlaga} first pretrains a structure-aware
multimodal aligner that combines visual and textual attributes through a
graph-guided objective. During instruction tuning, a lightweight projector
maps the resulting representations into the token space of a frozen LLM, with task-specific demonstrations, for node classification and link prediction. CHARM
\citep{yang2026charm} encodes retrieved structural, hierarchical, and
modality-complementary context into continuous graph tokens for zero-shot node classification and link prediction on unseen graph domains (\cref{sec:learn:mgfm}). 
GraphGPT-O
\citep{fang2025graphgpto} uses node-level and graph-level Q-Formers that first
fuse each neighbor's text and image tokens and then compress the neighborhood
into a fixed set of subgraph tokens. InstructG2I
\citep{jin2024instructg2i} uses a Graph-QFormer to produce a compact set of
graph prompts for image diffusion in G2Image (\cref{sec:fm:generation}).

\textbf{Structure-aware attention.}
Graph4MM \citep{ning2025graph4mm} follows a different route. Its
Hop-Diffused Attention restricts self-attention to graph edges and diffuses it
over multiple hops with decaying weights, while an MM-QFormer extracts
text-conditioned visual features for the language model. Instead of treating
topology as an independent modality appended to the input, this design lets
graph structure shape intra- and inter-modal attention.

\subsection{MLLMs as Graph Predictors}
\label{sec:fm:predictor}

Foundation models can also produce graph-level decisions directly, rather
than serving only as feature extractors for a downstream GNN. This includes
node classification and link prediction using multimodal node attributes,
selected graph context, and task instructions. The model may be prompted,
used with in-context examples, or fine-tuned for graph prediction.

\textbf{Roles of foundation models.}
The GraphVLM benchmark \citep{liu2026graphvlm} organizes the use of
vision--language models on MAGs into three roles: \emph{VLM-as-Encoder}, which
supplies multimodal features to graph learners; \emph{VLM-as-Aligner}, which
bridges modalities in a latent or linguistic space for LLM-based graph
reasoning; and \emph{VLM-as-Predictor}, which uses a VLM as the
decision-making backbone. The first two correspond to the feature and learning
levels (\cref{sec:feature,sec:learn}); the third places task prediction in the
foundation model. The simplest predictors fine-tune an off-the-shelf VLM such as LLaVA or Qwen-VL with LoRA for node classification, optionally adding serialized neighbor context to the prompt \citep{liu2026graphvlm}; MAGB additionally evaluates VLMs as zero-shot predictors \citep{yan2025magb}.

\textbf{Modality-adaptive prediction.}
Mario \citep{sun2026mario} targets node classification and link
prediction and adapts an LLM predictor to two properties of MAGs: the text and image of a node may be only weakly consistent, and
different nodes benefit from different modalities. It first trains a
graph-conditioned vision--language model in which a topology-aware mixer lets
the summary tokens of sampled nodes attend to one another with
shortest-path-distance biases under a text--image contrastive objective. It
then instruction-tunes an LLM with a router that selects, for each node, a
text-only, image-only, or multimodal prompt view.


\subsection{Graph-Conditioned Multimodal Generation}
\label{sec:fm:generation}

Most MAG benchmarks emphasize discriminative tasks. Graph-conditioned
generation instead uses a node's multimodal neighborhood to generate
missing textual or visual content, including G2Text and G2Image
(\cref{sec:prelim:tasks}). We focus on using a given graph to condition
content generation, rather than generating graph structures themselves.

\textbf{Problem formulation.}
Let $\mathcal{M}^{\star}\subseteq\mathcal{M}$ be the target set of modalities,
allowing either a single target (text or image) or joint text--image
generation. Denote the observed attributes of node $i$ by
$\mathbf{x}_i^{\mathrm{obs}}=\{x_i^m:m\in\mathcal{M}\setminus\mathcal{M}^{\star}\}$,
and let $\mathcal{G}_i^{-}$ be its ego-subgraph with the target attributes
of node $i$ withheld. A context selector $S$ returns a subgraph
$\mathcal{G}_i^{C}$ containing selected nodes, their attributes, and
relevant edges; $h_G$ converts this context into conditioning features
or tokens. A conditional generator $p_{\vartheta}$ then produces
\begin{equation}
\label{eq:gen}
\begin{aligned}
\mathcal{G}_i^{C}&=S\big(\mathcal{G}_i^{-},\mathbf{x}_i^{\mathrm{obs}}\big),\\
\hat{\mathbf{x}}_i^{\mathcal{M}^{\star}}&\sim
p_{\vartheta}\big(\,\cdot\mid\mathbf{x}_i^{\mathrm{obs}},
          h_G(\mathcal{G}_i^{C})\big).
\end{aligned}
\end{equation}
This separates \emph{context selection}, \emph{context encoding}, and the
\emph{generator and its control mechanism}, as summarized in
\cref{tab:generation}.

\begin{table*}[t]
\centering
\caption{Representative graph-conditioned generation approaches, organized
by the components of \cref{eq:gen}. PLANET reuses existing generation
pipelines to evaluate learned MAG representations.}
\label{tab:generation}
\footnotesize
\begin{tabularx}{\textwidth}{l l >{\raggedright\arraybackslash}X >{\raggedright\arraybackslash}X >{\raggedright\arraybackslash}X >{\raggedright\arraybackslash}X}
\toprule
Method & Output & Generator & Context selection & Graph interface & Control \\
\midrule
MMGL \citep{yoon2023mmgl} & Text & Pretrained LM & Section and page neighbors & Neighbor content and structure encodings & --- \\
Graph4MM \citep{ning2025graph4mm} & Text & Pretrained LM & Multi-hop graph neighbors & Hop-Diffused Attention and MM-QFormer & --- \\
InstructG2I \citep{jin2024instructg2i} & Image & Diffusion model & PPR and semantic re-ranking & Graph-QFormer tokens & Graph CFG \\
GraphGPT-O \citep{fang2025graphgpto} & Text, image & MLLM with diffusion decoder & PPR-based sampling & Linearization or hierarchical aligner & Generation order \\
PLANET \citep{liu2026toward} & Text, image & Existing pipelines & Inherited from pipeline & Topology-enriched node embeddings & Inherited \\
\bottomrule
\end{tabularx}
\end{table*}

\textbf{Context selection.}
Large multimodal neighborhoods are expensive to encode, particularly when
neighbor images produce many visual tokens. InstructG2I
\citep{jin2024instructg2i} selects neighbors using personalized PageRank
(PPR) followed by text--image semantic re-ranking. Let $\hat{\mathbf A}$
be a row-normalized adjacency matrix and $\beta\in(0,1)$ the propagation
coefficient. The PPR matrix satisfies
\begin{equation}
\label{eq:ppr}
\mathbf P=\beta\hat{\mathbf A}\mathbf P+(1-\beta)\mathbf I.
\end{equation}
For target node $i$, candidates with large $P_{ij}$ are re-ranked by
CLIP similarity between the target text and candidate images.
GraphGPT-O also uses PPR-based sampling. Such selection combines structural
proximity with semantic relevance; because it depends on both modalities, it
is itself a cross-modal channel, although a non-differentiable one
(\cref{sec:prelim:sensitivity}).

\textbf{Context encoding.}
The interfaces of \cref{sec:fm:sequence,sec:fm:softtokens} also serve
conditional generation (\cref{tab:generation}). PLANET
\citep{liu2026toward} provides a complementary evaluation setting: its learned
topology-enriched node embeddings replace input features in established MMGL
and InstructG2I pipelines, testing whether better graph representations also
improve generation.

\textbf{Generators and control.}
MMGL uses a pretrained language model for text, InstructG2I conditions a
diffusion backbone for image generation, and GraphGPT-O uses an MLLM with a
diffusion decoder for text--image generation
\citep{yoon2023mmgl,jin2024instructg2i,fang2025graphgpto}. InstructG2I further
introduces graph classifier-free guidance. Let $\epsilon_{\vartheta}$
be the denoising predictor, $\mathbf z_t$ a noisy latent, $c_T$ the text
condition, $c_G$ the graph condition, and $s_T,s_G\ge0$ guidance strengths.
Its guided prediction can be written as
\begin{equation}
\label{eq:graph-cfg}
\begin{aligned}
\hat\epsilon_{\vartheta}(\mathbf z_t,c_G,c_T)
={}&\epsilon_{\vartheta}(\mathbf z_t,\varnothing,\varnothing)\\
&+s_T\big[\epsilon_{\vartheta}(\mathbf z_t,\varnothing,c_T)
-\epsilon_{\vartheta}(\mathbf z_t,\varnothing,\varnothing)\big]\\
&+s_G\big[\epsilon_{\vartheta}(\mathbf z_t,c_G,c_T)
-\epsilon_{\vartheta}(\mathbf z_t,\varnothing,c_T)\big].
\end{aligned}
\end{equation}
Here $s_T$ and $s_G$ separately tune the influence of text and graph
context. Multiple graph conditions may be assigned distinct strengths,
allowing the contribution of different neighbor groups to be adjusted;
this is graph-conditioning control, not automatic independent control of
every edge \citep{jin2024instructg2i,ho2022cfg}.

\textbf{Evaluation.}
G2Text methods such as MMGL and Graph4MM report BLEU, ROUGE-L, and CIDEr on
Wikipedia section summarization \citep{yoon2023mmgl,ning2025graph4mm}. G2Image
methods compare generated and ground-truth images with CLIP and DINOv2
similarity and FID \citep{jin2024instructg2i}; GraphGPT-O additionally measures
text perplexity, image--text consistency, and agreement with the neighborhood
\citep{fang2025graphgpto}. Gains from the graph are usually shown against
text-only or general-purpose generators (e.g., Stable Diffusion, DreamLLM),
while PLANET reuses the MMGL and InstructG2I protocols to compare node
embeddings \citep{liu2026toward}. OpenMAG standardizes both tasks as downstream
probes \citep{wan2026openmag}; benchmark protocols are discussed in
\cref{sec:bench}.

\subsection{Analysis}
\label{sec:fm:analysis}

\textbf{When does structure help a foundation model?}
The GraphVLM benchmark shows that fine-tuned VLM predictors are
strong even without structural input and that the benefit of structure depends
on graph density and on how structure is supplied \citep{liu2026graphvlm}.
This does not make topology unnecessary: the relevant comparison is whether
structured context improves predictions beyond the same foundation model given
node attributes alone. Meaningful ablations should therefore hold the backbone,
node attributes, and training protocol fixed while changing only the structural
information supplied.


\textbf{Trade-offs among interfaces.}
Serializing neighbors keeps their content but lengthens the input: in MMGL, the
variant that feeds raw neighbor text performs best but requires much longer
inputs \citep{yoon2023mmgl}. Compressing the neighborhood into a fixed number
of tokens, as in the hierarchical aligner of GraphGPT-O
\citep{fang2025graphgpto}, shortens the input, but the relations among
neighbors must then be captured by the graph-side encoder. Structure-aware
attention keeps these relations explicit by restricting attention to graph
edges, at the cost of modifying the backbone's attention
\citep{ning2025graph4mm}.

\textbf{Connections across integration levels.}
Beyond serving directly as predictors and generators, foundation models also
enter the other integration levels. At the feature level, MLLMs such as LLaVA
and Qwen-VL supply node embeddings to GNNs and raise their performance
\citep{yan2025magb} (\cref{sec:feature}). Graph-centric designs remain
complementary on top of such embeddings: even with already aligned VLM features
(Qwen2-VL, LLaMA-3.2-Vision), ReCoG still yields
significant gains \citep{xue2026recog} (\cref{sec:prop:coupled}). At the
learning level, multimodal graph foundation models pretrain on frozen
foundation-model encoders, and their embeddings can in turn replace CLIP
features in generation pipelines \citep{he2025unigraph2,liu2026toward}
(\cref{sec:learn:mgfm}).
\section{Cross-Cutting Issues: Robustness, Efficiency, and Federation}
\label{sec:cross}

The integration of multimodal attributes and graph structure introduces
challenges that span all five levels of our taxonomy. We discuss three
cross-cutting issues: robustness to incomplete or corrupted information,
computational efficiency across different stages of the learning pipeline,
and federated learning under distributed multimodal graph data.

\subsection{Robustness}
\label{sec:cross:robust}

\textbf{Missing modalities and cold-start nodes.}
In practical applications, new nodes may lack certain modalities, graph
connections, or both. These cases are particularly challenging because
they limit the semantic and structural evidence available for prediction.
NTSFormer \citep{hu2026ntsformer} addresses isolated cold-start node
classification through a self-teaching Graph Transformer. A cold-start
attention mask enables the same model to produce a student prediction
using only the target node's attributes and a teacher prediction using
both self and neighborhood information. The teacher guides the student
without requiring a separate MLP, while multimodal precomputation and
mixture-of-experts input projections accommodate heterogeneous and
missing features.

In our formulation, missing modalities correspond to
$\mathcal{M}_i\subsetneq\mathcal{M}$, while isolated nodes satisfy
$\mathcal{N}(i)=\emptyset$. Their combination tests whether a model can
transfer knowledge from structurally contextualized training nodes to
nodes with incomplete information; NTSFormer reports results separately for
isolated nodes missing text, missing images, or neither.

\textbf{Noisy modalities and noisy edges.}
Multimodal graph models may encounter corrupted text, degraded images,
unreliable edges, and noisy supervision, which methods at different levels
address. At the topology level, the residual refiner of ReCoG weakens
unreliable observed edges \citep{xue2026recog} (\cref{sec:topo:refine}); at the
propagation level, RoleMAG assigns neighbors different roles and completes a
degraded modality of the target from the stronger modality of its neighbors
\citep{zuo2026rolemag} (\cref{sec:prop:coupled}); and at the learning level,
GraphMNL avoids propagating supervision from unreliable branches
\citep{wu2026multimodal} (\cref{sec:learn:ssl}). These mechanisms can also fail:
refinement based on feature similarity may introduce spurious edges when
multimodal features are noisy or misaligned \citep{xue2026recog}. OpenMAG
evaluates robustness by injecting modality noise, label noise, and structural
sparsity \citep{wan2026openmag}.

\textbf{Distribution shift.}
Beyond local corruption, the distribution of one or both modalities may differ
between training and test data, for example when products come from different
categories or images differ in color or background. OMG-NAS
\citep{cai2024omgnas} addresses this setting through multimodal graph neural
architecture search over GNN, fusion, and prediction operators. It learns
sample weights that decorrelate unimodal and cross-modal features, reducing
reliance on spurious correlations, and reuses the best weights across candidate
architectures as a warm start for the search.

\subsection{Efficiency}
\label{sec:cross:efficiency}

\textbf{Feature encoding versus graph learning.}
The computational cost of MAG learning depends strongly on whether
textual and visual encoders are frozen or trained end-to-end. With frozen
encoders, feature extraction is performed once and subsequent training
operates on cached embeddings. In contrast, updating large text or vision
encoders introduces repeated forward and backward passes that may
dominate the overall cost. OpenMAG \citep{wan2026openmag} supports both
static and trainable encoders.

\textbf{Propagation-decoupled pipelines.}
Decoupled architectures move parameter-free graph propagation into
a preprocessing stage and train downstream aggregation modules on
precomputed multi-hop representations. This eliminates repeated
edge-dependent propagation from the subsequent training epochs,
although preprocessing and feature storage still incur costs.
CAMPA \citep{su2026campa} systematically analyzes the efficiency
of decoupled multimodal GNNs and introduces cross-modal alignment
during precomputed propagation and trainable trajectory aggregation.
LION \citep{li2026lion} follows a similar
paradigm, using Clifford-based geometric operations to support
cross-modal interaction. These methods reduce repeated graph traversal,
but their end-to-end efficiency also depends on preprocessing,
cached representation size, and aggregation complexity
(\cref{tab:prop-complexity}).

\textbf{Sparse structure learning.}
Learning semantic relations between all node pairs requires
$O(N^2)$ affinity storage when the dense matrix is materialized.
Scalable approaches restrict the candidate space through sparse KNN graphs
\citep{zhang2021mining,zhou2023tale}, node--anchor approximations
\citep{zhu2026tmte}, or, in ReCoG, target-centric candidates within sampled
mini-batches followed by coarse-to-fine selection \citep{xue2026recog}
(\cref{sec:topo:refine}). These approximations reduce computation but may
miss useful long-range relations.

\textbf{Foundation-model interfaces.}
Serializing graph context lengthens the input of an LLM or MLLM, whereas
soft-token interfaces \citep{fan2026mlaga,yang2026charm} bound the number of
tokens but still require graph sampling and encoding
(\cref{sec:fm:analysis}).

\subsection{Federated Multimodal Graph Learning}
\label{sec:cross:federated}

Multimodal attributed graphs are often distributed across organizations
or platforms that cannot directly share raw text, images, or relational
data. Federated multimodal graph learning enables collaborative training
under these constraints, but clients may differ simultaneously in modality
availability, feature distributions, graph structures, and label spaces.

\textbf{Benchmarking federated MAG learning.}
MM-OpenFGL \citep{li2026mm} formalizes multimodal federated
graph learning and provides a benchmark comprising 19 datasets across
7 application domains, 8 simulation strategies, 57 implemented methods,
and 6 downstream tasks. Its simulated federated settings capture
differences in modality availability, label distributions, and
topological information, enabling systematic evaluation of multimodal
learning under heterogeneous clients.

\textbf{Federated pretraining and alignment.}
FedGAMMA \citep{li2026toward} extends multimodal graph foundation
learning to federated settings (\cref{sec:learn:mgfm}). It combines
shared--private semantic enhancement with optimal-transport-based
cross-modal alignment, topology-aware graph fusion, and affinity-aware
aggregation across clients. The pretrained model is subsequently adapted
through lightweight graph-aware prompts, reducing the need for full
model fine-tuning at each client.
\section{Benchmarks, Dataset Characteristics, and Evaluation Protocols}
\label{sec:bench}

This section reviews the benchmark platforms published in recent years, their
datasets and the metrics used for each task, model-independent dataset
descriptors, and the evaluation protocols together with their baselines.
Because the platforms differ in datasets, encoders, splits, and baselines, we
compare their designs rather than pooling accuracies from incompatible
experiments.

\subsection{Benchmark Platforms}
\label{sec:bench:platforms}

\cref{tab:benchmarks} summarizes five representative benchmark platforms.

\textbf{MM-GRAPH.}
MM-GRAPH \citep{zhu2025mmgraph} provides seven text--image-attributed datasets
of varying scale covering node classification (NC), link prediction (LP), and
knowledge-graph completion (KGC). It evaluates standardized GNN, multimodal
GNN, and multimodal KG baselines under aligned and unaligned feature encoders
(CLIP, ImageBind, and T5 with ViT or DINOv2). KGC datasets are listed for
completeness, although relational KG algorithms are outside the main scope of
this survey.

\textbf{MAGB.}
MAGB \citep{yan2025magb} curates five graphs spanning Amazon product networks
(Movies, Toys, Grocery) and social media (Reddit-S, Reddit-M). It studies
GNN-as-Predictor and zero-shot VLM-as-Predictor paradigms, with an emphasis on
modality preference and the choice of feature encoder, and finds that textual
attributes are more informative in e-commerce graphs and visual attributes in
social graphs.

\textbf{GraphVLM.}
GraphVLM \citep{liu2026graphvlm} evaluates VLM-as-Encoder, -Aligner, and
-Predictor paradigms for node classification on six Amazon and Reddit datasets,
including cross-dataset transfer.

\textbf{OpenMAG.}
OpenMAG \citep{wan2026openmag} integrates 19 datasets across six domains,
16 encoders supporting frozen and trainable extraction, and 24 models in a
unified library. Its eight tasks span graph-centric evaluation (NC, LP,
clustering) and modality-centric evaluation (matching, retrieval, alignment,
G2Text, G2Image), and it also tests robustness to modality noise, label noise,
and structural sparsity.

\textbf{MM-OpenFGL.}
MM-OpenFGL \citep{li2026mm} extends benchmarking to federated MAG
learning with 19 datasets across seven domains, eight simulation strategies
that vary modality availability and topology across clients, 57 methods, and
six downstream tasks (\cref{sec:cross:federated}).

\begin{table*}[t]
\centering
\caption{Representative MAG benchmarks. Task abbreviations: NC, node
classification; LP, link prediction; KGC, knowledge-graph completion;
Clust, node clustering; MM, modality matching; MR, modality retrieval;
MA, modality alignment; G2T/G2I, graph-conditioned text/image generation.}
\label{tab:benchmarks}
\footnotesize
\setlength{\tabcolsep}{4pt}
\begin{tabularx}{\textwidth}{@{}l l c >{\raggedright\arraybackslash}X >{\raggedright\arraybackslash}X >{\raggedright\arraybackslash}X@{}}
\toprule
Platform & Venue & \#Datasets & Domains / encoders & Main tasks & Compared paradigms \\
\midrule
MM-GRAPH \citep{zhu2025mmgraph} & CVPR'25 & 7 & Products, books, MMKG; CLIP, ImageBind, T5, ViT, DINOv2 & NC, LP, KGC & MLP, GCN, SAGE, BUDDY; MMGCN, MGAT; MoSE, VISTA \\
\addlinespace
MAGB \citep{yan2025magb} & KDD'25 & 5 & E-commerce and social media; text, image, and VLM features & NC; zero-shot category prediction & GNN-as-Predictor; VLM-as-Predictor \\
\addlinespace
GraphVLM \citep{liu2026graphvlm} & CVPR'26 & 6 & Amazon products and Reddit; CLIP-style and VLM features & NC, including cross-dataset transfer & VLM-as-Encoder / Aligner / Predictor \\
\addlinespace
OpenMAG \citep{wan2026openmag} & ICML'26 & 19 & Six domains; 16 static/trainable encoders & NC, LP, Clust; MM, MR, MA, G2T, G2I & 24 GNN, MAG, and FM-enhanced models \\
\addlinespace
MM-OpenFGL \citep{li2026mm} & arXiv'26 & 19 & Seven domains; heterogeneous client modalities and topologies & Six downstream tasks (federated) & 57 methods; eight federation simulations \\
\bottomrule
\end{tabularx}
\end{table*}

\subsection{Datasets}
\label{sec:bench:datasets}

\cref{tab:datasets} lists representative datasets with the statistics reported
by their source benchmark or by the indicated study, together with the metrics
typically reported for each task. Edge counts follow the convention of each
source (directed arcs or undirected pairs), and ranking metrics for LP and KGC
depend on the candidate set and negative-sampling policy, so numbers are not
directly comparable across sources.

\begin{table*}[t]
\centering
\caption{Representative multimodal graph datasets with the statistics, tasks,
and metrics reported by their sources. \#Cls: number of classes (NC datasets
only). KGC edge counts are positive relation-labeled triples summed over the
training, validation, and test splits. Goodreads-NC uses the MM-GRAPH
configuration also adopted by ReCoG. A listed task identifies a reported use,
not every task supported by the platform. Metrics: Acc, accuracy; MRR, mean
reciprocal rank; Hits@$K$, fraction of positives ranked in the top $K$; NMI,
normalized mutual information; ARI, adjusted Rand index; CLIP/DINOv2, embedding similarity between generated and reference
images
\citep{zhu2025mmgraph,yan2025magb,wan2026openmag,jin2024instructg2i,xue2026recog}.}
\label{tab:datasets}
\footnotesize
\setlength{\tabcolsep}{4pt}
\begin{tabularx}{\textwidth}{@{}lllrrcl>{\raggedright\arraybackslash}X@{}}
\toprule
Dataset & Source / version & Domain & $|\mathcal V|$ & $|\mathcal E|$ & \#Cls & Tasks & Metrics \\
\midrule
Ele-fashion & MM-GRAPH & Products & 97,766 & 199,602 & 12 & NC & Acc, F1 \\
Goodreads-NC & MM-GRAPH & Books & 685,294 & 7,235,084 & 10 & NC & Acc, F1 \\
Amazon-Sports & MM-GRAPH & Products & 50,250 & 356,202 & -- & LP & MRR, Hits@$K$ \\
Amazon-Cloth & MM-GRAPH & Products & 125,839 & 951,271 & -- & LP & MRR, Hits@$K$ \\
Goodreads-LP & MM-GRAPH & Books & 636,502 & 3,437,017 & -- & LP & MRR, Hits@$K$ \\
MM-CoDEx-s & MM-GRAPH & MMKG & 1,383 & 15,884 & -- & KGC & MRR, Hits@$K$ \\
MM-CoDEx-m & MM-GRAPH & MMKG & 7,697 & 52,840 & -- & KGC & MRR, Hits@$K$ \\
\midrule
Movies & MAGB / OpenMAG & Products & 16,672 & 218,390 & 20 & NC & Acc, F1 \\
Toys & MAGB / OpenMAG & Products & 20,695 & 126,886 & 18 & NC, Clust & Acc, F1; NMI, ARI \\
Grocery & MAGB / OpenMAG & Products & 17,074 & 171,340 & 20 & NC & Acc, F1 \\
Reddit-S & MAGB / OpenMAG & Social & 15,894 & 566,160 & 20 & NC, Clust & Acc, F1; NMI, ARI \\
Reddit-M & MAGB & Social & 99,638 & 1,167,188 & 50 & NC & Acc, F1 \\
\midrule
DY & OpenMAG & Video & 8,299 & 35,627 & -- & LP & MRR, Hits@$K$ \\
KU & OpenMAG & Video & 5,370 & 22,052 & -- & LP & MRR, Hits@$K$ \\
Bili\_dance & OpenMAG & Video & 2,307 & 9,127 & -- & LP & MRR, Hits@$K$ \\
Flickr30k & OpenMAG & Image--text & 31,783 & 181,151 & -- & G2Text & BLEU, ROUGE-L, CIDEr \\
SemArt & OpenMAG & Art & 21,382 & 1,216,432 & -- & G2Image & CLIP, DINOv2 \\
\midrule
ART500K & InstructG2I & Art & 311,288 & 643,008,344 & -- & G2Image & CLIP, DINOv2, FID \\
Amazon (Beauty) & InstructG2I & Products & 178,890 & 3,131,949 & -- & G2Image & CLIP, DINOv2, FID \\
Goodreads (Mystery) & InstructG2I & Books & 93,475 & 637,210 & -- & G2Image & CLIP, DINOv2, FID \\
\bottomrule
\end{tabularx}
\end{table*}

\subsection{Dataset Characteristics}
\label{sec:bench:characteristics}

Dataset size alone does not explain why different integration mechanisms
help. We therefore characterize MAG datasets with descriptors computed from
graph structure and labels only, so that they depend neither on a feature
encoder nor on a downstream predictor (\cref{tab:characteristics}).
Throughout, $y_i$ denotes the label of node $i$.

\textbf{Label homophily ($h_y$).}
Label homophily is the fraction of edges whose endpoints share the same class.
Let $\mathcal E_L\subseteq\mathcal E$ contain the edges whose endpoints are
both labeled, and let $\mathbb{I}[\cdot]$ be the indicator function; then
\begin{equation}
\label{eq:bench-homophily}
h_y=\frac{1}{|\mathcal E_L|}\sum_{(i,j)\in\mathcal E_L}
\mathbb{I}[y_i=y_j].
\end{equation}
It is defined only for datasets with node labels; for multi-label datasets,
the indicator is replaced by a label-set similarity such as the Jaccard index.
Label homophily measures how well the observed graph agrees with the prediction
target: a low value indicates that propagation along observed edges mixes
classes, which motivates the topology-level refinement and heterophily-aware
designs of \cref{sec:topo}.

\textbf{Average degree ($\bar d$).}
The average degree is the mean number of neighbors per node. Let $d_i$ be the
degree of node $i$ in the undirected graph without self-loops or duplicate
edges; then
\begin{equation}
\label{eq:bench-degree}
\bar d=\frac{1}{|\mathcal V|}\sum_{i\in\mathcal V}d_i=\frac{2|\mathcal E|}{|\mathcal V|}.
\end{equation}
Computing it on the deduplicated graph makes it comparable across datasets,
whereas raw edge counts follow different conventions (\cref{sec:bench:datasets}).
It indicates how much neighborhood evidence propagation can use: the benefit
of structure for VLM predictors is largest on dense graphs and unstable on
sparse ones \citep{liu2026graphvlm}, and sparse graphs leave more room for
recovering missing edges through topology-level methods (\cref{sec:topo}).

\textbf{Clustering coefficient (CC) and transitivity (Tr).}
Both measure how often the neighbors of a node are themselves connected. Let
$d_i$ be the degree of node $i$ and $T_i$ the number of triangles that contain
$i$; the local clustering coefficient is $c_i=2T_i/[d_i(d_i-1)]$ if $d_i\ge2$
and $c_i=0$ otherwise. The average clustering coefficient and the transitivity
are
\begin{equation}
\label{eq:bench-cc}
\begin{aligned}
\mathrm{CC}&=\frac{1}{|\mathcal V|}\sum_{i\in\mathcal V}c_i,\\
\mathrm{Tr}&=\frac{3\times\#\text{triangles}}{\#\text{connected triples}}.
\end{aligned}
\end{equation}
CC averages local closure over nodes, whereas Tr measures it globally and is
dominated by high-degree nodes. They distinguish graphs whose neighborhoods
form dense local communities from sparse, tree-like graphs, which differ in
how much information additional propagation hops contribute. The two can differ
markedly: on Ele-fashion, CC (0.173) is about three times Tr (0.056), indicating
that the neighbors of high-degree nodes, which dominate Tr, are rarely
connected to one another, whereas the Reddit graphs reach $\mathrm{Tr}=1$
because they consist of cliques (see edge provenance below).


\begin{table}[t]
\centering
\caption{Label homophily ($h_y$), average degree ($\bar d$), average clustering
coefficient (CC), and transitivity (Tr); the upper block lists NC datasets and
the lower block LP datasets. Values for the MM-GRAPH datasets are reported by
\citet{zhu2025mmgraph}; values for the MAGB datasets are computed on the
released graphs from distinct undirected edges with the same definitions,
which reproduce the MM-GRAPH values on Ele-fashion. ``N/A'': not applicable for
datasets without node labels.}
\label{tab:characteristics}
\footnotesize
\setlength{\tabcolsep}{3pt}
\begin{tabular}{@{}l l c c c c@{}}
\toprule
Dataset & Source & $h_y$ & $\bar d$ & CC & Tr \\
\midrule
Movies & MAGB & 0.4470 & 9.65 & 0.3442 & 0.2584 \\
Toys & MAGB & 0.7241 & 5.48 & 0.2398 & 0.1908 \\
Grocery & MAGB & 0.6813 & 8.33 & 0.3071 & 0.1445 \\
Reddit-S & MAGB & 0.9585 & 17.81 & 0.6993 & 1.0000 \\
Reddit-M & MAGB & 0.6716 & 5.86 & 0.5762 & 1.0000 \\
Ele-fashion & MM-GRAPH & 0.7675 & 4.08 & 0.1730 & 0.0560 \\
Goodreads-NC & MM-GRAPH & 0.6663 & 21.11 & 0.1614 & 0.0498 \\
\midrule
Amazon-Sports & MM-GRAPH & N/A & 14.18 & 0.4002 & 0.2658 \\
Amazon-Cloth & MM-GRAPH & N/A & 15.12 & 0.2940 & 0.1846 \\
Goodreads-LP & MM-GRAPH & N/A & 10.79 & 0.1102 & 0.0348 \\
\bottomrule
\end{tabular}
\end{table}

\textbf{Edge provenance.}
Beyond these descriptors, the process that generates edges determines what a
graph encodes: co-purchase or co-view relations in product graphs, interactions
in Reddit, and shared artists or genres in art graphs. In MAGB's Reddit
graphs, two posts are linked when the same user commented on both
\citep{yan2025magb}; their transitivity of 1.0000 (\cref{tab:characteristics})
indicates that these graphs consist almost entirely of cliques. The same raw
data can yield different graphs: OpenMAG builds Flickr30k from grounding
annotations that link image regions and caption phrases \citep{wan2026openmag}, whereas
PLANET builds a feature-similarity $k$NN graph over Flickr30k
\citep{liu2026toward}. When edges are built from node features, the topology
and the features carry overlapping evidence.


\subsection{The Baseline Landscape: Protocols and Their Baselines}
\label{sec:bench:baselines}

An evaluation protocol specifies the graph version, the encoder and its
training status, the data split, the task, and the metric. A platform can
support several protocols, and a later method that uses the same dataset need
not follow the original protocol. \cref{tab:protocols} maps each platform
setting, together with the method-specific configuration of ReCoG, which
combines datasets from two platforms, to its encoders, splits, and baseline
families.

\begin{table*}[t]
\centering
\caption{Representative evaluation settings and baseline families.
``Method-specific'' denotes an experimental configuration rather than an
independent benchmark. Dataset overlap does not guarantee identical splits,
encoders, or evaluators.}
\label{tab:protocols}
\footnotesize
\setlength{\tabcolsep}{4pt}
\begin{tabularx}{\textwidth}{@{}l >{\raggedright\arraybackslash}X >{\raggedright\arraybackslash}X >{\raggedright\arraybackslash}X >{\raggedright\arraybackslash}X@{}}
\toprule
Setting & Datasets / task & Encoders and partitions & Baseline families & Representative use \\
\midrule
MM-GRAPH & Ele-fashion, Goodreads-NC (NC); Sports, Cloth, Goodreads-LP (LP); MM-CoDEx-s/m (KGC) & CLIP, ImageBind, T5+ViT, T5+DINOv2; NC 60/10/30; Amazon LP 80/10/10; Goodreads LP 60/10/30 & Feature-only MLP; GCN, SAGE, BUDDY (LP); MMGCN, MGAT; MoSE, VISTA (KGC) & MM-GRAPH \citep{zhu2025mmgraph}; ReCoG \citep{xue2026recog} \\
\addlinespace
MAGB & Movies, Toys, Grocery, Reddit-S, Reddit-M (NC) & Text, image, and VLM embeddings; released splits & MLP and conventional GNNs; zero-shot VLM predictor & MAGB \citep{yan2025magb}; NSG-MoE \citep{zhang2026nsgmoe}; ReCoG \citep{xue2026recog} \\
\addlinespace
OpenMAG & 19 datasets; 3 graph and 5 modality tasks & 16 frozen/trainable encoders; 60/20/20 & MLP; GNNs; graph-enhanced, multimodal-enhanced, and FM-enhanced methods & OpenMAG \citep{wan2026openmag} \\
\addlinespace
GraphVLM & Movies, Toys, Grocery, Arts, CDs, Reddit (NC) & CLIP-style and VLM encoders; 60/20/20; zero-shot or fine-tuned & GNNs on VLM features; graph-LLM aligners; VLM predictors & GraphVLM \citep{liu2026graphvlm} \\
\addlinespace
ReCoG (method-specific) & MM-GRAPH NC/LP plus MAGB NC (nine datasets) & Four MM-GRAPH and three MAGB encoder configurations & MLP, GCN, SAGE, MMGCN, MGAT, NSG-MoE, LION, PLANET, UniGraph2 & ReCoG \citep{xue2026recog} \\
\bottomrule
\end{tabularx}
\end{table*}
\section{Applications}
\label{sec:apps}

\textbf{E-commerce and recommendation.}
Product graphs associate titles, reviews, and images with co-purchase or
co-view relations, supporting product categorization, link prediction, and
recommendation \citep{zhu2025mmgraph,yan2025magb}. Early multimodal GNNs
were developed for recommendation over multimedia user--item graphs
\citep{wei2019mmgcn,tao2020mgat}, followed by methods that construct
modality-specific item relations or capture global semantic dependencies
\citep{zhang2021mining,zhou2023tale,guo2024lgmrec,hu2025miggt}. Newly
added items often lack both edges and some modalities; NTSFormer
\citep{hu2026ntsformer} classifies such isolated cold-start nodes with
incomplete multimodal features (\cref{sec:cross:robust}).

\textbf{Social media and content platforms.}
Social media graphs connect text--image posts through discussion threads or
comments by the same user, while short-video graphs link videos that are
co-viewed or co-liked and associate them with visual covers and textual
descriptions \citep{yan2025magb,wan2026openmag}. These settings support
content classification, link prediction, and recommendation, where graph
context provides relational information beyond individual posts or videos.

\textbf{Art and image--text graphs.}
Art graphs connect artworks through shared artists or genres, and
graph-conditioned image generation uses these relations to synthesize
content consistent with related artistic styles
\citep{jin2024instructg2i,fang2025graphgpto}. Image--text graphs link image
regions and caption phrases; OpenMAG uses such a graph built from Flickr30k
for graph-conditioned text generation and cross-modal retrieval
\citep{wan2026openmag}.

\textbf{Structured web content and documents.}
Encyclopedic pages contain sections, images, and captions whose relations
extend beyond isolated image--text pairs. MMGL \citep{yoon2023mmgl} and
Graph4MM \citep{ning2025graph4mm} exploit these relations for
graph-conditioned text generation, in particular Wikipedia section
summarization.

\textbf{Other domains and scope.}
MAG learning also extends to industrial applications: C-MAG \citep{li2025c}
predicts which manufacturers can make which products by propagating
multimodal manufacturer attributes and images through a two-stage cascade of
graphs. Multimodal knowledge graphs and healthcare graphs are further
application areas, but their specialized problem formulations are outside
the scope of this survey \citep{chen2024kgmm,vaida2026healthcare}.
\section{Multimodal Graph Learning and AI Agents}
\label{sec:ge}

AI agents increasingly use graphs to organize perception, memory, planning,
tool execution, and multi-agent coordination \citep{bei2025graphsagents}.
As agents process text, images, code, and other forms of information,
graph-based representations provide opportunities to connect multimodal
learning with agent systems. However, not all agent graphs are multimodal
attributed graphs (MAGs). We distinguish two settings. The first concerns
graphs that agents perceive, construct, or remember, which can be modeled as
MAGs when nodes retain multiple perceptual or semantic modalities
(\cref{sec:ge:perceive}). The second concerns workflow and communication
graphs that organize agent computation (\cref{sec:ge:engineering}). These
graphs are not necessarily MAGs, but their information-flow mechanisms
suggest connections to multimodal graph learning.

We use \emph{graph engineering} primarily to refer to the design of
workflow and communication structures. This is narrower than broader
formulations that also include runtime state and memory management
\citep{grapheng2026}. Rather than treating these settings as equivalent,
we examine which principles of multimodal graph learning may transfer
to agent systems and outline directions for future research.

\subsection{Multimodal Graphs for Agent Perception and Memory}
\label{sec:ge:perceive}

Multimodal agents perceive visual environments, screens, videos, and textual
instructions, and may organize their observations into structured graphs.
Embodied agents use scene graphs to represent objects and spatial
relationships \citep{rana2023sayplan,yin2024sgnav}, which can be enriched
with multimodal semantic attributes \citep{song2024scenemmkg}. GUI agents
can represent navigation and interface transitions as graphs
\citep{sun2025guixplore}. Similarly, agent memory systems organize
experiences and retrieved evidence through graph structures
\citep{yang2025damcs,liu2025memverse}. VimRAG
\citep{wang2026vimrag}, for example, constructs a dynamic memory graph
and allocates visual tokens according to the topological roles of
memory nodes.

When graph nodes retain multiple modalities, these structures can be
modeled as MAGs, making the methods reviewed in this survey potentially
applicable. Topology refinement may suppress unreliable associations
(\cref{sec:topo}), graph-aware alignment may connect visual observations
with textual descriptions (\cref{sec:learn:align}), and missing-modality
learning may support incomplete memories (\cref{sec:cross:robust}).
Foundation-model interfaces can further convert structured multimodal
memories into representations accessible to LLMs and MLLMs
(\cref{sec:fm}).

Nevertheless, agent-generated graphs differ from conventional MAG
benchmarks in several important respects. Their topology may evolve
during interaction, node attributes may be incomplete or revised, and
errors in graph construction can influence subsequent retrieval and
reasoning. Moreover, their usefulness is often determined by downstream
task success rather than node classification accuracy. Although existing
MAG benchmarks examine missing modalities and robustness, the joint
evaluation of dynamically constructed multimodal graphs and agent-level
decision quality remains insufficiently explored.

\subsection{Graph Engineering and Cross-Channel Interaction}
\label{sec:ge:engineering}

Graph engineering organizes the execution and communication of agent
systems. A workflow graph
$\mathcal{W}=(\mathcal{V}_{\mathcal{W}},\mathcal{E}_{\mathcal{W}})$
contains operators, such as LLM calls, tools, agents, and validators,
as nodes, with directed edges representing execution dependencies or
information transfer. Unlike MAGs, workflow nodes do not necessarily
represent entities with multiple semantic modalities. Nevertheless,
workflow representations and multi-agent communication exhibit several
structural principles related to multimodal graph learning.

\textbf{Multi-view workflow representations.}
\label{sec:ge:multiview}
FLORA \citep{zhang2025flora} predicts workflow performance using GNNs
over prompt-derived node features. Agentic Predictor
\citep{agenticpredictor2025} extends this direction by representing
workflow operators through multiple views, including prompts, code,
and operator information. It propagates each view over a shared
workflow topology, combines the resulting representations through
cross-view attention, and incorporates workflow-level information
for performance prediction.

This architecture resembles modality-decoupled propagation followed by
fusion (\cref{eq:decoupled}). Under fixed topology and independent
per-view propagation, the representation of one view does not depend
on features from other views before fusion, as characterized by
\cref{obs:jacobian}. However, workflow views should not automatically
be identified with perceptual modalities, and the analogy does not
imply that their information is statistically or semantically equivalent.

This connection motivates examining whether coupled cross-view
propagation, view-dependent graph structures, and reliability-aware
fusion can improve workflow representations. For example, code
dependencies and prompt-level semantic dependencies may require
different neighborhood structures, analogous to modality-specific
topologies in MAG learning (\cref{sec:topo:hetero}).

\textbf{Communication topology.}
Multi-agent systems also use graphs to determine which agents exchange
information. G-Designer \citep{zhang2025gdesigner} constructs
task-adaptive communication graphs, while AgentPrune
\citep{zhang2025agentprune} prunes redundant communication in
multi-agent workflows. These designs are related to graph structure
learning in that they determine which connections should carry
information (\cref{eq:unified-structure}).

However, communication edges may also impose execution constraints,
incur token and latency costs, or determine whether information reaches
downstream agents. Consequently, topology refinement in agent systems
must consider not only semantic relevance but also communication cost,
dependency preservation, and task-level utility.

\textbf{Connections to multimodal information flow.}
\label{sec:ge:analogy}
During execution, agents exchange natural-language reasoning,
structured tool results, code, visual observations, and other forms
of evidence. We refer to these as \emph{information channels}.
Although channels are not necessarily modalities in the MAG sense,
several multimodal graph learning principles provide useful analogies
when multiple agents or tools process evidence about a shared target. \cref{tab:ge-analogy} summarizes these connections. They are intended
as potential design principles rather than established equivalences
between multimodal graph learning and agent systems.

\begin{table}[t]
\centering
\caption{Connections between multimodal graph learning and
agent information flow. The analogies suggest possible designs
rather than established theoretical equivalences.}
\label{tab:ge-analogy}
\footnotesize
\begin{tabularx}{\columnwidth}{
>{\raggedright\arraybackslash}p{0.32\columnwidth}
>{\raggedright\arraybackslash}X}
\toprule
MGL principle & Potential connection to agent systems \\
\midrule
Fusion timing &
Independent agents followed by aggregation (late fusion) versus
intermediate information exchange during reasoning. \\
\addlinespace
Cross-modal coupling &
Evidence from one channel can influence the processing of
other channels before the final decision. \\
\addlinespace
Graph-aware alignment &
Grounding different channels in shared entities, evidence,
or subgoals to maintain semantic consistency. \\
\addlinespace
Role-aware propagation &
Routing information according to source reliability,
complementarity, or verification needs
\citep{zuo2026rolemag}. \\
\addlinespace
Shared--private learning &
Preserving independent evidence while extracting common
conclusions across agents. \\
\addlinespace
Over-smoothing and residuals &
Avoiding premature consensus by retaining independent
initial reasoning across communication rounds. \\
\bottomrule
\end{tabularx}
\end{table}

A key distinction concerns cross-channel sensitivity. The Jacobian
analysis of \cref{obs:jacobian} describes differentiable dependencies
between modality representations. Agent execution instead often
involves discrete messages, sampled outputs, and non-differentiable
tool calls. Therefore, cross-channel influence should be evaluated
through controlled interventions rather than directly transferring
the Jacobian results.

Let $\mathcal{K}$ be the set of information channels, $y^k$ the
observed content of channel $k$, $\tilde y^k$ a counterfactual
version, and $\mathcal{D}$ a query distribution. Let $a(q;\operatorname{do}(y^k\leftarrow v),\xi)$ denote
the final output for query $q$ when channel $k$ is assigned content
$v$ at a specified stage, with downstream execution rerun and
randomness represented by $\xi$. An operational measure of
cross-channel decision sensitivity is
\begin{equation}
\label{eq:ge-influence}
\begin{aligned}
I_k = \Pr_{q\sim\mathcal{D},\,\xi}\big[
&a(q;\operatorname{do}(y^k\leftarrow y^k),\xi)\\
&\neq
a(q;\operatorname{do}(y^k\leftarrow\tilde y^k),\xi)
\big].
\end{aligned}
\end{equation}
This measures whether changing a channel affects the final decision, but not whether its contribution is beneficial. To assess utility, let
$U(a,q)$ be a task-specific reward or success measure and define
\begin{equation}
\label{eq:ge-utility}
\begin{aligned}
\Delta U_k = \mathbb{E}_{q,\xi}\big[
&U(a(q;\operatorname{do}(y^k\leftarrow y^k),\xi),q)\\
-&U(a(q;\operatorname{do}(y^k\leftarrow\tilde y^k),\xi),q)
\big].
\end{aligned}
\end{equation}
A paired intervention should hold the upstream execution state and
other experimental conditions fixed where possible, while rerunning
steps that depend on the modified channel. Measuring both $I_k$ and
$\Delta U_k$ distinguishes sensitivity to information changes from
their actual contribution to task success.

\textbf{Limits of the analogy.}
Unlike modality representations, agent messages are discrete,
stochastic, and generated through state-dependent computation.
Communication topology affects execution dependencies as well as
information flow, and modifying an edge may change whether certain
tools or agents are reachable. Furthermore, many workflows decompose
a task into distinct subtasks whose nodes do not describe a common
entity. In such cases, heterogeneous workflow modeling may be more
appropriate than a multimodal node representation. These distinctions
limit direct transfers of MAG theory but motivate adapting its
structural principles to agent-specific constraints.

\subsection{Future Research Directions}
\label{sec:ge:hypotheses}

The connections above suggest several directions for investigating
how multimodal graph learning principles can inform agent systems.
We emphasize them as open research questions rather than established
performance claims.

\textbf{Dynamic multimodal perception and memory graphs.}
Agent memories and perceptual graphs evolve as new observations arrive
and previously stored evidence is revised. Future work could investigate
modality-aware topology refinement, missing-modality recovery, and
reliability-aware retrieval in dynamically constructed graphs. A central
question is whether these mechanisms improve agent task success under
incomplete or corrupted multimodal observations, rather than merely
improving static graph prediction metrics.

\textbf{Cross-view workflow representation learning.}
Existing multi-view workflow predictors motivate studying interactions
between different workflow views before final fusion. Coupled
cross-view propagation and view-dependent topology learning may capture
dependencies overlooked by independent view encoders. Their benefits
should be tested against strong late-fusion baselines under matched
workflow representations, model capacity, and training protocols,
including settings with missing or unreliable views.

\textbf{Adaptive and cost-aware communication.}
Agent communication graphs could incorporate ideas from topology
refinement and role-aware message passing to select information
exchange according to source reliability, complementarity, and
downstream requirements. Unlike conventional MAG learning, these
decisions must also account for token consumption, latency, and
execution dependencies. An important direction is to determine when
additional communication improves task utility enough to justify
its computational cost.

\textbf{Preserving independent reasoning.}
Repeated communication may cause agent outputs to become increasingly
similar, potentially suppressing useful independent evidence.
Shared--private representation learning and residual propagation
suggest mechanisms for maintaining both consensus and diversity.
Future studies could investigate whether preserving initial
reasoning or selectively exchanging complementary evidence
improves multi-round collaboration. Controlled channel
interventions, such as \cref{eq:ge-influence,eq:ge-utility},
could help distinguish genuine information exchange from
superficial answer agreement.

Overall, multimodal graph learning and agent graph engineering
address different computational settings but share structural
questions about how heterogeneous information should be
represented, propagated, aligned, and selectively preserved.
Understanding which principles transfer, and under what conditions, offers a promising direction for connecting multimodal graph algorithms with increasingly structured AI agent systems.

\section{Open Problems and Future Directions}
\label{sec:open}

Despite recent progress, multimodal graph learning faces several unresolved
questions concerning expressiveness, generalization, scalability, and
evaluation. We distinguish open scientific problems (\cref{sec:open:problems}) and concrete
community efforts that could advance the field in the near term
(\cref{sec:open:efforts}), and then outline research directions suggested by
recent trends (\cref{sec:open:directions}).

\subsection{Open Problems}
\label{sec:open:problems}

\textbf{Expressiveness theory of cross-modal message passing.}
\cref{obs:jacobian} distinguishes architectures whose intermediate
representations are sensitive to other modalities from those whose
modalities remain decoupled before fusion. Such sensitivity is necessary
for expressing cross-modal dependencies at the propagation stage, but
does not by itself establish greater expressive power or downstream
benefits. Open questions include which functions over MAGs require
cross-modal coupling during propagation rather than at final fusion;
how coupled multimodal message passing relates to heterogeneous-graph
message passing through modality-specific node splitting
\citep{zhang2026nsgmoe}; and how topology-level cross-modal interactions
compare with propagation-level coupling. A Weisfeiler--Lehman-style
expressiveness hierarchy for MAGs could provide a theoretical basis
for comparing these architectures.

\textbf{When does structure help in the era of strong MLLMs?}
Fine-tuned vision--language models can perform strongly even without
explicit graph structure, and the benefit of incorporating topology is
largest on dense graphs and unstable on sparse ones \citep{liu2026graphvlm}.
Nevertheless, graph-aware interfaces can provide additional structural
information \citep{ning2025graph4mm, xue2026recog}. A central question is when relational
information provides benefits beyond strong pretrained multimodal
representations. Ideally, these benefits could be predicted from measurable
graph characteristics, such as label homophily, average degree, clustering,
and edge provenance (\cref{sec:bench:characteristics}), together with
encoder-dependent properties such as modality-wise neighborhood similarity and
disagreement. Understanding their interaction with task requirements and
encoder capacity would move the field beyond case-by-case empirical
comparisons.

\textbf{Unifying discriminative and generative tasks.}
Graph-centric and modality-centric tasks may favor different
representations. For example, TMTE employs fused representations for
graph-centric tasks and modality-specific representations for
modality-centric tasks \citep{zhu2026tmte}. Developing a transferable
backbone that supports node classification, link prediction,
cross-modal retrieval, and graph-conditioned generation remains
an important challenge. Such a model should preserve structural
and modality-specific information while supporting different tasks
through lightweight heads, prompts, or adapters, rather than
requiring full backbone retraining for each task. This direction
is closely related to the broader goal of multimodal graph
foundation models and MG-LLMs \citep{wang2025mgllm}.

\textbf{Efficiency at billion scale.}
Industrial graphs may contain billions of nodes, making multimodal
encoding, graph propagation, and topology learning potential
computational bottlenecks. Their relative costs depend on the
training and deployment setting: frozen and precomputed modality
embeddings reduce repeated encoder computation, whereas online
encoding or fine-tuning large multimodal models may dominate
the total cost. Meanwhile, learning or refreshing graph structure
can introduce expensive candidate-search and similarity-computation
operations. Open directions include encoder-aware sampling,
incremental embedding caching and refresh, sparse topology
refinement, and budget-aware selection of which modality
representations to compute.

\textbf{Dynamic multimodal graphs.}
The benchmarks reviewed in this survey are predominantly static,
yet real-world products, posts, videos, and their relationships
evolve continuously. Nodes and edges may arrive or disappear,
while textual and visual attributes can change independently.
Open problems include temporal multimodal message passing,
incremental topology refinement, efficient representation updates,
and robustness to stale or asynchronous modalities. Temporal MAG
benchmarks with realistic update patterns and evaluation protocols
are needed to study these challenges systematically.

\textbf{Beyond text and images.}
Current MAG benchmarks predominantly emphasize text and images,
while many applications involve additional modalities such as
video, audio, tabular records, and sensor measurements. These
settings introduce heterogeneous temporal resolutions, modality
availability patterns, and encoding costs. They also raise
questions about whether different modalities require distinct
receptive fields, propagation operators, or neighborhood
structures. Architectures with pairwise cross-modal interactions, such as CAMPA
\citep{su2026campa}, may incur quadratic costs in the number of modalities
(\cref{tab:prop-complexity}), whereas coupling through scalar coefficients and
linear aggregation, as in ReCoG \citep{xue2026recog}, can merge intra- and
cross-modal aggregation (\cref{sec:prop:coupled}). This motivates sparse,
hierarchical, or selectively coupled designs that remain effective as the
modality set expands.

\textbf{Trustworthiness.}
Reliability remains insufficiently understood beyond standard
prediction accuracy. Missing or corrupted modalities, noisy
edges, and shifts in graph structure can affect different
integration levels in distinct ways (\cref{sec:cross:robust}).
Federated MAG learning further introduces privacy and
cross-client heterogeneity concerns \citep{li2026mm}.
Another open question is how to explain predictions in terms
of both contributing modalities and influential graph
relationships. Future models should be evaluated not only
for predictive accuracy but also for robustness, privacy,
calibration, and explanation fidelity under controlled
perturbations.

\textbf{Multimodal graph learning for AI agents.}
Perception and memory graphs of agents can be modeled as MAGs when nodes retain
multiple modalities, whereas workflow and communication graphs share only
related information-flow challenges. Whether the methods of this survey improve
agent reasoning and coordination remains largely untested (\cref{sec:ge}).

\subsection{Concrete Efforts for the Community}
\label{sec:open:efforts}

Addressing these problems requires not only new architectures
but also more informative datasets, reproducible evaluation,
and shared research infrastructure. We highlight six
concrete efforts.

\textbf{E1: Characterize every dataset.}
Benchmarks should release systematic descriptors alongside
each dataset, following \cref{sec:bench:characteristics}.
Encoder-independent statistics should include graph scale,
connectivity, clustering, label homophily, edge semantics,
structural provenance, and modality availability before
imputation. Encoder-dependent descriptors, such as
modality-wise edge similarity and modality disagreement,
should be computed using specified frozen reference
encoders. These descriptors should be distinguished from
downstream model performance. Standardized dataset
characterization would make it possible to study how
structural and modality properties relate to architectural
choices without conflating dataset characteristics with
model effectiveness.

\textbf{E2: A versioned, shared evaluation protocol.}
A shared protocol should specify dataset versions,
preprocessing, feature encoders, official splits, random
seeds, and task-specific evaluation procedures. Results
should be reported across multiple seeds and, where
appropriate, additional splits. For link prediction,
held-out evaluation edges must be excluded from
training-time message-passing structure to prevent
information leakage. Strong non-graph baselines,
including feature-only predictors using the same
modality embeddings and MLLMs without graph structure,
are essential for isolating the contribution of topology.
Controlled text-only, image-only, and multimodal
ablations should quantify task-dependent modality
complementarity. Finally, evaluations should report
end-to-end runtime, peak memory, and preprocessing
costs (\cref{sec:bench:baselines,sec:cross:efficiency}).

\textbf{E3: Controlled synthetic benchmarks.}
Synthetic MAGs with known dependencies between modalities,
nodes, labels, and graph topology could provide controlled
tests of architectural expressiveness. For example, labels
could depend on cross-modal interactions within a node,
cross-modal information from neighbors, or relationships
that require topology refinement. SMGFM uses a small controlled synthetic MAG as a mechanism test for text--image retrieval \citep{wu2026smgfm}; broader benchmark families could
systematically vary label homophily, modality disagreement,
missingness, and neighbor roles
\citep{zuo2026rolemag}. Such settings would allow
early-fusion, modality-decoupled, propagation-coupled,
and topology-coupled models to be compared under
controlled conditions, helping separate representational
limitations from optimization and dataset effects.

\textbf{E4: Scalable joint topology and representation learning.}
Methods that jointly optimize graph topology and
multimodal representations \citep{zhu2026tmte,xue2026recog}
illustrate the potential benefits of adapting relational
structure to modality information. However, candidate
edge construction, pairwise scoring, and repeated
structure updates may become prohibitively expensive
on large graphs. Developing sparse, incremental, and
mini-batch-compatible refinement mechanisms for graphs
with $10^7$ nodes or more is therefore an important
practical direction. Such approaches should avoid dense
global similarity matrices and support efficient updates
when nodes, edges, or modality embeddings change.

\textbf{E5: Open multimodal graph foundation models.}
The community would benefit from openly released
multimodal graph foundation models with reproducible
pretraining procedures, documented data provenance,
and standardized transfer protocols. These models
should support both graph-centric and modality-centric
tasks, ideally through lightweight task-specific
interfaces. Releasing pretrained checkpoints,
feature encoders, evaluation code, and clear zero-shot
and few-shot settings would help determine whether
multimodal graph pretraining provides transferable
advantages over task-specific GNNs and strong
pretrained multimodal encoders
(\cref{sec:learn:mgfm,sec:fm:analysis}).

\textbf{E6: Realistic missingness and perturbations.}
Dataset releases should preserve original modality availability and
pre-imputation information, so that natural missingness can be distinguished
from simulated corruption. Controlled evaluations should further
examine attribute noise, edge perturbations, distribution shifts, and
heterogeneous modality quality (\cref{sec:cross:robust}).

\subsection{Future Research Directions}
\label{sec:open:directions}

The five integration levels reviewed in this survey suggest
several emerging directions that go beyond improving individual
architectural components. In particular, future multimodal graph
models may increasingly learn how modalities and structure
should interact, rather than relying on predetermined
integration mechanisms.

\textbf{Adaptive integration across levels.}
Recent methods demonstrate the benefits of combining multiple
integration levels. ReCoG jointly learns refined topology and
multimodal representations through coupled propagation, while
TMTE iteratively optimizes topology and modality representations
for downstream tasks \citep{xue2026recog,zhu2026tmte}.
A promising direction is to make such integration adaptive
rather than applying a fixed strategy to all nodes and
modalities. Future models could determine when cross-modal
interaction is necessary, which neighbors should exchange
information across modalities, and whether such interaction
should occur through feature fusion, propagation, or topology
refinement. These decisions could depend on modality
reliability, neighborhood characteristics, and task
requirements, balancing complementary information,
modality conflict, and computational cost.

\textbf{Graph-native multimodal foundation models.}
Recent foundation-model interfaces increasingly incorporate
graph structure into multimodal processing. Graph4MM introduces
Hop-Diffused Attention to integrate multi-hop topology into
attention \citep{ning2025graph4mm}, while Mario employs
modality-adaptive routing to select informative graph-aware
instruction views for each node and its neighborhood
\citep{sun2026mario}. CHARM constructs hierarchical graph
contexts for transferable reasoning \citep{yang2026charm},
and InstructG2I uses graph-based context selection to
condition image generation \citep{jin2024instructg2i}.
These methods motivate a further step toward graph-native
MLLMs, in which structural reasoning is integrated into
the model's attention and computation rather than provided
only through externally encoded graph tokens. An important
direction is to jointly learn task-dependent neighborhood
retrieval, modality selection, and structural attention,
allowing the model to determine which graph evidence
is needed for a given query while controlling context
length and computational cost.

\textbf{Closing the loop between generation and representation.}
Discriminative graph representations can already improve
multimodal generation pipelines, as demonstrated by
topology-aware representations evaluated through
graph-conditioned generation \citep{liu2026toward}.
The reverse direction remains less explored: generated
modalities could supplement incomplete node attributes,
extending reconstruction-based pretraining
(\cref{sec:learn:mgfm}) from embedding space to raw content. This suggests a unified learning framework in which graph
representations guide generation and generated evidence, when reliable, improves representation learning and downstream prediction.
Key challenges include preventing synthetic attributes
from reinforcing model errors, estimating generation
uncertainty, and determining when generated information
provides useful evidence beyond existing modalities.

\textbf{Learning graph construction from raw multimodal data.}
Most MAG methods assume an observed graph or refine
connections within a predefined candidate space
(\cref{sec:topo}). However, the construction of nodes
and relationships from raw multimodal observations is
itself an important modeling problem. Related systems
have demonstrated agent-based construction of multimodal
knowledge graphs from unstructured evidence
\citep{park2026m}, although such knowledge graphs
fall outside the primary scope of this survey.
For MAG learning, a promising direction is to jointly
infer relational candidates from raw text and images,
estimate their reliability, and refine the resulting
graph using downstream task feedback. Such systems
would move beyond reweighting an existing topology
toward learning how the graph should be constructed
in the first place, while accounting for uncertain
relations, noisy evidence, and the cost of graph updates.

\textbf{Higher-order multimodal interactions.}
Most multimodal graph architectures model interactions
through pairwise edges and node-level representations.
LGMRec demonstrates how hypergraphs can capture global
dependencies among groups of users and items
\citep{guo2024lgmrec}, while Clifford-based methods
such as LION and ICE introduce algebraic representations
that preserve richer interaction components
\citep{li2026lion,li2026ice}. Although hypergraph
relations and higher-grade algebraic interactions
represent different notions of higher-order structure,
both suggest alternatives to conventional pairwise
message passing. Future work could investigate
modality-aware hyperedges, group-level cross-modal
propagation, and structured latent interaction spaces
that capture dependencies not adequately represented
by pairwise aggregation. Understanding their
expressiveness and computational trade-offs would
be particularly important for scaling these models
to large multimodal graphs.

\section{Conclusion}
\label{sec:conclusion}

This survey examined multimodal graph learning through the question of
where and how multimodal information and graph structure are integrated.
We introduced a unified message-passing perspective and organized
representative methods into five integration levels: feature, propagation,
topology, learning, and foundation-model. This framework provides a
common basis for comparing discriminative and graph-conditioned
generative approaches, highlighting their architectural assumptions,
information-flow mechanisms, and computational trade-offs.

We also reviewed five recent benchmarking efforts, mapped their evaluation
protocols to the baselines they use, and characterized representative
datasets with encoder-independent descriptors. Beyond existing MAG
applications, we explored potential connections to AI agents and graph
engineering, and identified open scientific problems, emerging algorithmic
directions, and concrete community efforts. We hope this survey will provide a systematic foundation for the community to
better understand existing methods, design more informative evaluations, and
develop more advanced architectures that determine when, where, and how
multimodal information and graph structure should interact.

\bibliography{refs}
\bibliographystyle{icml2026}

\end{document}